%% file: arxiv.tex
\documentclass{article}
\usepackage{amssymb}
\usepackage{amsmath}
\usepackage{tabularx}
\usepackage{booktabs} 
\usepackage{siunitx}
\usepackage{multirow}
\usepackage{graphicx}
\usepackage{makecell} 
\usepackage{enumitem}
\usepackage{wrapfig}
\usepackage{float}
\usepackage[hang,flushmargin]{footmisc} 
\usepackage[preprint]{corl_2025} 

\title{Generalist Robot Manipulation\\beyond Action Labeled Data}

\author{
    Alexander Spiridonov$^{1, 2, \ast}$,
    Jan-Nico Zaech$^{1}$,
    Nikolay Nikolov$^{1}$,\\
    \textbf{Luc Van Gool$^{1}$,
    Danda Pani Paudel$^{1}$}
    \\
    \normalsize{$^{1}$INSAIT, Sofia University “St. Kliment Ohridski”, Bulgaria}\\ 
    \normalsize{$^{2}$ETH Zurich, Switzerland}\\
    \normalsize{$^\ast$Corresponding author: \href{mailto:alexander-marc.spiridonov@insait.ai}{\tt alexander-marc.spiridonov@insait.ai}}\\
}

\begin{document}
\maketitle
\vspace{-3.0em}
\begin{center}
\url{https://motovla.github.io/}
\end{center}

\begin{figure}[h]
    \centering
    \includegraphics[width=1.0\columnwidth]{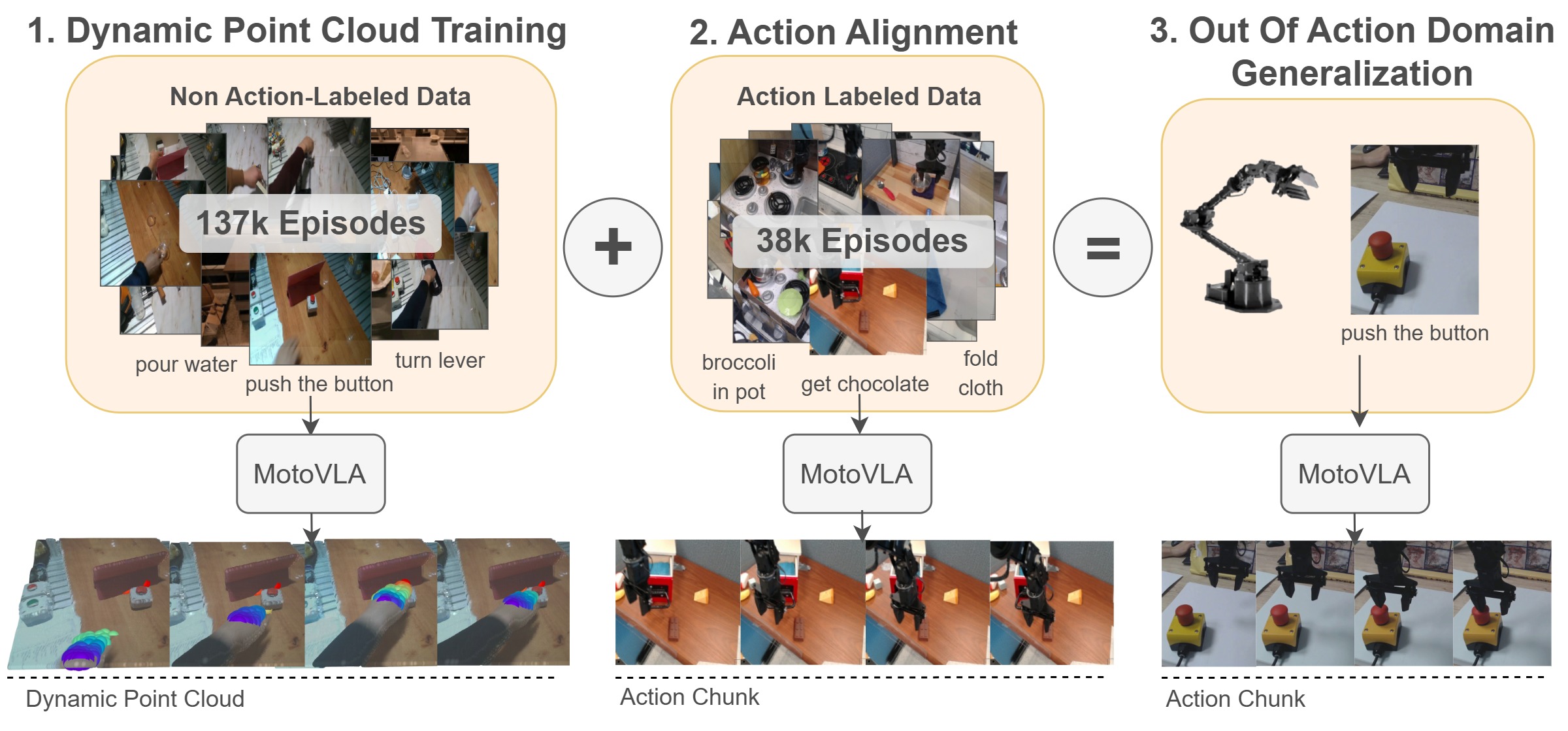}
    \caption{Our method enables robot actions whose labels are not available during training. Such unlabeled demonstrations may come from humans (e.g. ``push the button") or other robots performing them. This out-of-action domain generalization is achieved using the proposed large-scale dynamic point cloud-based training, followed by the action alignment on a smaller dataset with action labels.}
    \label{fig:intro}
\end{figure}
\label{subsec:architecture}

\begin{abstract}
Recent advances in generalist robot manipulation leverage pre-trained Vision–Language Models (VLMs) and large-scale robot demonstrations to tackle diverse tasks in a zero-shot manner. A key challenge remains: scaling high-quality, action-labeled robot demonstration data, which existing methods rely on for robustness and generalization. To address this, we propose a method that benefits from videos without action labels—featuring humans and/or robots in action—enhancing open-vocabulary performance and enabling data-efficient learning of new tasks. Our method extracts dense, dynamic 3D point clouds at the hand or gripper location and uses a proposed 3D dynamics predictor for self-supervision. This predictor is then tuned to an action predictor using a smaller labeled dataset for action alignment. We show that our method not only learns from unlabeled human and robot demonstrations—improving downstream generalist robot policies—but also enables robots to learn new tasks without action labels (i.e., out-of-action generalization) in both real-world and simulated settings.  
\end{abstract}



\section{Introduction}

Robust zero-shot manipulation across diverse tasks and environments is one of the biggest bottlenecks towards truly autonomous robots. Inspired by the open-world reasoning capabilities of Large-Language (LLM) and Vision-Language Models (VLM), Vision-Language-Action (VLA) models have emerged for generalist robot manipulation. Approaching this challenge, VLAs extend the semantic reasoning abilities of VLMs with embodied understanding and adapt them for robotic control by training on large datasets of teleoperated robot demonstrations~\cite{OpenVLA,pi0,oxe,cotvla,cogact,revla,spatialvla}. This has led to impressive progress in learning robust manipulation policies. However, most success is centered around in-domain settings, and performance quickly degrades as the tasks move outside the training distribution. While collecting yet larger robot datasets seems straightforward, it remains unclear what resources would be required to achieve generalist manipulation.

Multimodal training with videos of human demonstrations is a promising alternative to prohibitively expensive robot demonstrations. Such videos contain valuable spatiotemporal information highly relevant to learning robotic control, are readily available at internet scale, and provide diverse tasks and environments. However, learning from human demonstration datasets comes with a range of challenges; Videos provide no direct action labels supervision, exhibit human-to-robot domain gaps, and include redundant or distracting features irrelevant to robotic control.

Being a fundamental challenge, learning motion priors from humans and unlabeled data has been widely explored. Yet existing work remains confined to specialist, small-scale policies. Some focus on visual representations~\cite{R3M,Masked_Visual_Pre-training,VIP,Where}, not considering unseen motions. Others predict visual plans that require a bespoke inverse-dynamics model for execution~\cite{UniPi,Gen2Act,VILP,Flow2Act,ATM,Track2Act}. Another line retargets human hands to robot grippers~\cite{PointPolicy,MotionTracks,EgoMimic,ZeroShot,HandObject}, but suffers from a large domain gap.

In this work, we bridge this gap and present MotoVLA, a generalist robot manipulation policy that enables new tasks from human and robot videos without action labels. To achieve this, we propose a VLA model and two-stage training approach using a combination of large-scale labeled and unlabeled
\footnote{\noindent  Unlabeled refers to \textit{non-action-labeled}, as action labels are the main challenge in acquiring manipulation data.}
human and robot videos.
In the first training stage, a dynamic point cloud predictor is trained on the unlabeled data, which establishes a common embodiment-agnostic action representation. Since the dynamic point cloud strongly correlates with the end-effector actions up to hand-eye calibration, the second stage training of an action expert on action-labeled data is simplified. This natural correspondence between dynamic point clouds and 3D robot actions makes our approach particularly effective for learning from unlabeled data. An overview of our method is shown in Figure~\ref{fig:intro}.
In summary, our contributions are:
\begin{itemize}
    \vspace{-5pt}
  \setlength{\itemsep}{0pt}
  \setlength{\parskip}{0pt}
  \setlength{\parsep}{0pt}
    \item MotoVLA, the first end-to-end VLA model that allows the use of unlabeled data for learning motion priors required for the generalist robot manipulation.
    \item A two-stage training approach enabling the use of dynamic point clouds as a common embodiment-agnostic representation, which is both scalable and intuitive.
    \item Extensive real and simulated evaluations of our model for in-domain, out-of-domain, and transfer learning tasks, demonstrating the effective use of unlabeled data by our model.
\end{itemize}        

\section{Related Work}
\label{sec:related_work}

\textbf{Generalist Robot Manipulation} relies on zero-shot capabilities, which have been demonstrated in vision and language through internet-scale training. To enable this progress in robotics, two dominant paradigms have formed:
The first group follows a modular approach and utilizes pre-trained models to generate high-level plans. In vision space, language-conditioned generative models are used to create visual manipulation plans, executed by inverse dynamics policies~\cite{Gen2Act,UniPi,susie,genima}. 
Similarly, the semantic reasoning capabilities of VLMs~\cite{pali3,Paligemma,gemma2,SigLIP, VLM} are utilized for tasks decomposition and high-level planning~\cite{SayCan,huang2022language,ProgPrompt,Palme} in language space. While this utilizes pretrained generative models, performance is limited by the need for accurate inverse models with low-level zero-shot capabilities.

The second group is VLAs, end-to-end models~\cite{OpenVLA,pi0,oxe,cotvla,cogact,revla,spatialvla} that implicitly perform planning, state estimation, and subtask execution by extending VLMs to robotic control. These models process vision and language inputs and directly predict robot commands. As the first open-source model, OpenVLA~\cite{OpenVLA} builds on top of the Prismatic VLM~\cite{Prismatic}. $\pi_0$~\cite{pi0} introduces a flow-matching-based action expert into VLAs and builds on top of Paligemma~\cite{Paligemma}. These VLAs demonstrate impressive zero-shot language-conditioned capabilities for tasks and environments close to their training distribution, while out-of-domain zero-shot performance quickly degrades. Recently, some works have aimed to improve VLA generalization by including non-action-labeled video data during training~\cite{3DVLA, cotvla}. However, they are only capable of predicting goal states, thereby missing the motion information present in the full video trajectories. 
To the best of our knowledge, our approach is the first to learn fine-grained motion priors from large-scale unlabeled demonstrations directly with an end-to-end generalist VLA architecture.

\textbf{Transfer Learning of Robotic Manipulation} from video offers a scalable alternative to teleoperated robot data. Transfer is typically achieved by utilizing intermediate representations, either implicitly or explicitly:
Using \textit{implicit visual representations}~\cite{He_2022_CVPR}, downstream policy learning~\cite{R3M,Masked_Visual_Pre-training,VIP,Where} has demonstrated visual generalization; however it has yet to show generalization to new actions. 
Direct transfer of motion has been achieved by \textit{explicit mapping} of hand movements from human demonstrations to robot actions through retargeting~\cite{PointPolicy,MotionTracks,EgoMimic,ZeroShot,HandObject} with key-points and estimated hand poses.
Other explicit representations predict video or optical flow for a given task~\cite{UniPi,Gen2Act,VILP,Flow2Act,ATM,Track2Act}, but require subsequent imitation learning using action-paired demonstrations and fall behind non-hierarchical models.

Closest to our work, recent non-hierarchical approaches~\cite{cotvla,Arm4r,LAPA} extract latent actions, 3D scene flow, or sub-goal images from unlabeled demonstrations and use them as part of the training mixture~\cite{cotvla,Arm4r,LAPA}.
However, these policies have not been successfully scaled to generalist VLA policies, and all of these works investigate small sets of in-domain tasks.

In contrast, our work uses dense hand and gripper point cloud sequences as a representation, which are interpretable, abstract the embodiment, and directly encode the spatial and temporal relationships of the underlying motions. This enables the transfer of human demonstration in a generalist zero-shot VLA setting, for both in-domain and out-of-domain tasks.          

\section{Method}
\label{sec:method}

In the following, Sec.~\ref{subsec:prelim} and Sec.~\ref{subsec:data_annotation} introduce the different data sources and the dynamic point cloud representation. Sec.~\ref{subsec:architecture} details the VLA architecture, and Sec.~\ref{subsec:training} describes the two-stage training approach to integrate labeled and unlabeled data.

\subsection{Training Objectives and Data.}
\label{subsec:prelim}

\textbf{Learning from Unlabeled Videos.} We denote the action-free stage~one training dataset as $\mathcal{T}_o = \{(\tau_o^{(i)}, l^{(i)})\}_{i=1}^{N_o}$, where $l^{(i)}$ is the language description of the $i^{th}$ episode and $\tau_o^{(i)} = \{\mathbf{I}_t^{(i)}\}_{t=1}^T$ is the corresponding sequence of camera RGB images. To encode the spatio-temporal information, we propose dynamic point clouds, which, for episode $i$, is a temporal sequence of point clouds denoted as $\mathbf{P}^{(i)} = \{\mathbf{p}_t^{(i)}\}_{t=1}^T$. Each point cloud $\mathbf{p}_t^{(i)} \in \mathbb{R}_{n \times 3}$ contains $n$ consistent points representing the hand or robot gripper in the frame of the camera.
This data is used for stage~one self-supervised video pre-training. The objective is the prediction of the future dynamic point cloud of a given trajectory, conditioned on the past point cloud. Thus, our VLA learns the mapping
\begin{equation}
    \mathbf{f}_\theta^{points}(\mathbf{I}_t^{(i)}, l^{(i)},\mathbf{p}_{t-h:t}^{(i)}) \to \mathbf{p}_{t:t+c}^{(i)}.
\end{equation}
For better readability we denote $\mathbf{P}_t = \mathbf{p}_{t:t+c}$.

\textbf{Learning from Robot Demonstrations.} Although the dynamic point clouds capture the motion dynamics of the hand and robot gripper, a direct or model-based conversion to robot commands is not possible in a generalizable manner. Instead, we assume that there is a potentially considerably smaller action-labeled dataset, which we denote as $\mathcal{T}_a = \{(\tau_a^{(i)}, l^{(i)})\}_{i=1}^{N_a}$, where $\tau_a^{(i)} = \{\mathbf{I}_t^{(i)}, \mathbf{q}_t^{(i)}\}_{t=1}^T$ is a sequence of RGB images and robot proprioceptive observations.
We then use this dataset to further train the VLA to directly predict robot actions in the corresponding coordinate frame, corresponding to the mapping
\begin{equation}
    \mathbf{f}_\theta^{act}(\mathbf{I}_t^{(i)}, l^{(i)},\mathbf{q}_{t-h:t}^{(i)}) \to \mathbf{q}_{t:t+c}^{(i)},
\end{equation}
with the predicted future proprioception $\mathbf{q}_{t:t+c}^{(i)}$ being executable on the robot. This adaptation is possible with little training data, since it primarily aligns the dynamic point clouds to robot commands by performing implicit hand-eye calibration.
For better readability we denote $\mathbf{A}_t = \mathbf{q}_{t:t+c}$. 

\subsection{Dynamic Point Clouds from Unlabeled Videos}  
\label{subsec:data_annotation}

Dynamic hand and gripper point clouds provide a physically grounded supervision signal from unlabeled videos of humans or robots. To extract the dynamic point clouds from video data, we first sample a frame index $t \sim \mathcal{N}(T/2, 0.4 \cdot T)$ from the video. The hand or gripper is detected using Grounding DINO~\cite{GroundingDINO} by obtaining a bounding box, which is then passed to Segment Anything V2~\cite{SAM2} to generate a segmentation mask. We uniformly sample pixel positions from this mask and track them forward and backward in time using BootsTAPIR~\cite{BootsTAP}. We lift the resulting 2D tracks into 3D using the affine invariant monocular depth estimator MoGE~\cite{MoGE}, leveraging its predicted depth and camera intrinsics to produce a sequence of hand and gripper point clouds.

\begin{figure}
    \centering
    \includegraphics[width=0.7\columnwidth ,trim={5mm 5mm 0mm 5mm}, clip]{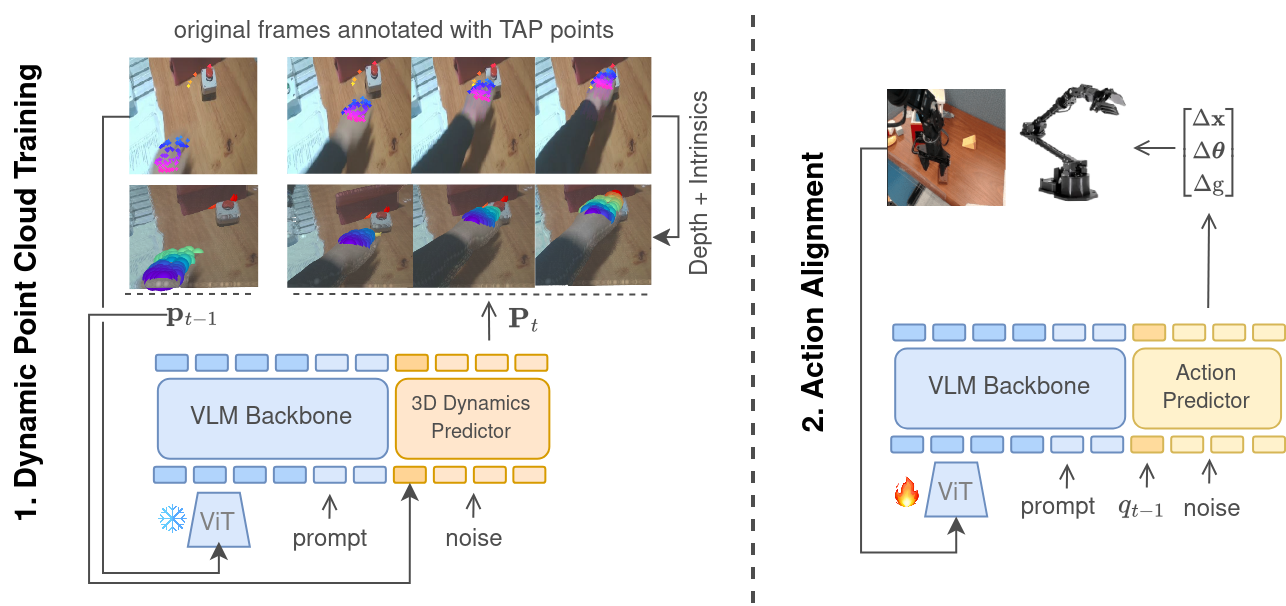}
    \caption{We utilize the 3D Dynamics Predictor to learn an embodiment agnostic representation from unlabeled videos (left), followed by the Action Predictor for generalist manipulation (right).}
    \label{fig:architecture}
    \vspace{-4mm}
\end{figure}

\subsection{VLA Architecture}
\label{subsec:architecture}

We use a Mixture of Transformers architecture~\cite{MixtureOfTransformers} during both the dynamic point cloud training and action alignment stages. The backbone of our VLA is a pre-trained VLM, enabling vision-language reasoning. A second smaller Transformer~\cite{Transformer} is used as \textit{Predictor} and attends to the VLM backbone. This Predictor is trained as \textit{3D Dynamics Predictor} during the first training stage and subsequently tuned as \textit{Action Predictor} during action alignment in the second training stage. The Predictor is conditioned on the VLM hidden features via self-attention over the concatenated keys and values and performs prediction using flow matching~\cite{FlowMatching}. For a complete forward pass, the VLM processes the image and text inputs and returns its key-value cache to the Predictor, where it is reused for each iteration of denoising. This decoupling allows the individual sets of weights to specialize in semantic and embodied reasoning in the VLM and Predictor, respectively, and speeds up inference.

\textbf{VLM Backbone.} Our VLA uses Paligemma~\cite{Paligemma} as the pretrained vison-language backbone. Paligemma is a 3 billion parameter model using the SigLIP~\cite{SigLIP} vision encoder and the Gemma~\cite{gemma} language model. The model is pre-trained on a diverse corpus of multimodal data for tasks such as vision question answering, captioning, and multi-object instance segmentation~\cite{Paligemma}. We chose Paligemma since it offers a trade-off between inference speed and semantic reasoning performance suitable for real-time robotic control tasks. During inference and training, the VLM Backbone processes the current image and text: $\mathbf{I}_{t}, l_t$ and returns the key-value cache $\mathbf{h_t}$ to the Predictor.

\textbf{3D Dynamics Predictor.} During the first training stage, the Predictor serves as a \textit{3D Dynamics Predictor}. Input to the model is the current hand or robot gripper point cloud $\mathbf{p}_{t-1}$, the noisy sample, and the key-value cache $\mathbf{h_t}$ as conditioning. By using flow matching, it predicts the future dynamic point cloud as a sequence of future point maps $\mathbf{P}_t$.

The \textit{3D Dynamics Predictor} utilizes conditional flow matching~\cite{FlowMatching} to sample from the continous distribution of future point maps $\mathbf{p}_{t:t+c} \sim p\left(\mathbf{p}_{t:t+c} \vert \mathbf{o}_t\right)$, where $\mathbf{o}_t = \left[\mathbf{h_t}, \mathbf{p}_{t-1} \right]$.
Accordingly, the dynamic point cloud prediction loss is 
\begin{equation}
	\mathcal{L}_{points} = \mathbb{E}_{p\left(\mathbf{P}_t \vert \mathbf{o}_t\right), \epsilon, \tau}\|\mathbf{v}_\theta^{points}\left(\mathbf{P}_t^\tau, \mathbf{o}_t\right) - \left(\textbf{P}_t - \left(1 - \sigma_{min}\right)\epsilon\right)\|^2, 
\end{equation}
where $\mathbf{P}_t^\tau  = \left(1 - \left(1 - \sigma_{min}\right)\tau\right)\epsilon + \tau \mathbf{P}_t$ is a noisy sample of $\mathbf{P}_t$ and $\epsilon\sim \mathcal{N}\left(\mathbf{0},\mathbf{I}\right)$. We use the time sampling, as introduced by $\pi_0$~\cite{pi0}, namely $\tau = \left(1 - \sigma_{min}\right)\left(1 - z\right)$, with $z \sim \text{Beta}(1.5, 1)$.

The \textit{3D Dynamics Predictor} model $\mathbf{v}_\theta^{points}\left(\mathbf{P}_t^\tau, \mathbf{o}_t\right)$ is implemented as a $300$ million parameter model, following the Gemma language model architecture. It projects the current point map $\mathbf{p}_{t-1}$ and noisy future dynamic point cloud $\mathbf{P}_t^\tau$ to the language model feature dimension via two separate linear projectors. The model outputs are decoded to dynamic point clouds via a linear projector. More details about the model architecture are provided in the Appendix~\ref{app:architecture}.

\textbf{Action Predictor.}
The \textit{Action Predictor} mirrors the \textit{3D Dynamics Predictor}. The model input is the current proprioception $\mathbf{q}_{t-1}$, the noisy sample, and the key-value cache $\mathbf{h_t}$ as conditioning. It predicts the next action chunk $\mathbf{A}_t$ using the same conditional flow-matching formulation. For the \textit{Action Predictor}, the observations is $\mathbf{o}_t = \left[\mathbf{h_t}, \mathbf{q}_{t-1} \right]$ and the loss is defined as
\begin{equation}
	\mathcal{L}_{action} = \mathbb{E}_{p\left(\mathbf{A}_t \vert \mathbf{o}_t\right), \epsilon, \tau}\|\mathbf{v}_\theta^{act}\left(\mathbf{A}_t^\tau, \mathbf{o}_t\right) - \left(\textbf{A}_t - \left(1 - \sigma_{min}\right)\epsilon\right)\|^2, 
\end{equation}
where $\mathbf{A}_t^\tau  = \left(1 - \left(1 - \sigma_{min}\right)\tau\right)\epsilon + \tau \mathbf{A}_t$ is a noisy sample of $\mathbf{A}_t$ and $\epsilon\sim \mathcal{N}\left(\mathbf{0},\mathbf{I}\right)$. The time $\tau$ distribution remains unchanged. The model is identical and initialized with the weights from the \textit{3D Dynamics Predictor} $\mathbf{v}_\theta^{points}$, while linear encoders and decoders are initialized randomly.

\subsection{Training}
\label{subsec:training}

We train MotoVLA in two phases. During the first phase, the \textit{3D Dynamics Predictor} is trained to generate the future dynamic point cloud given the current hand or gripper point cloud. This objective enables the model to jointly capture the underlying dynamics of both human and robot manipulation sequences. The subsequent second stage aligns the VLA to predict executable robot actions by training the initialized \textit{Action Predictor} on a subset of action-labeled demonstrations. A full overview of our two-stage training approach can be found in Figure~\ref{fig:architecture}.

\textbf{Dynamic Point Cloud Training Stage}. During the dynamic point cloud training, we use both human and robot demonstrations. For human demonstrations, we use the RH20T~\cite{rh20t} dataset, and for robot demonstrations BridgeData V2~\cite{BridgeV2} and Rt-1~\cite{brohan2022rt}. Although the robot demonstration datasets contain action labels, they are not utilized and discarded in the first stage. By predicting dynamic point clouds from both human and robot samples, the model is encouraged to learn a shared representation across the two domains. The VLM is initialized from a pre-trained checkpoint while the \textit{3D Dynamics Predictor} is randomly initialized. We train the VLM and \textit{3D Dynamics Predictor} on the $\mathcal{L}_{points}$ objective, but freeze the vision encoder to preserve the open-vocabulary visual capabilities of the VLM.

\textbf{Action Alignment Stage}. In the second training stage, the goal is to align the representations learned from dynamic point cloud training with robot action prediction. Since we evaluate on the WidowX robot, actions from the BridgeData V2~\cite{BridgeV2} are utilized. The VLM and \textit{Action Predictor} are initialized with weights from the first training stage, while the linear layers of the action encoder and decoder are initialized randomly. In the second training stage, the vision encoders are trainable, as this can improve action accuracy~\cite{OpenVLA}, and general vision capabilities are less influenced due to the shorter training. The entire model is trained using the action prediction objective $\mathcal{L}_{action}$.


\section{Experimental Results}
\label{sec:related_work}
The following section presents experiments to validate and investigate the effectiveness of the proposed method. Specifically, we aim to answer the following questions: 
\begin{enumerate}
  \setlength{\itemsep}{0pt}
  \setlength{\parskip}{0pt}
  \setlength{\parsep}{0pt}
    \item Does our method lead to better overall performance for in-domain tasks and environments?
    \item Does dynamic point cloud training transfer capabilities from unlabeled cross-embodiment demonstrations to new robot actions?
    \item How do different design choices surrounding the model architecture, training recipe, and action-less data labeling influence model performance?   
\end{enumerate} 
\vspace{-3mm}
\subsection{Implementation Details}
 
We train our model on a TPUv5e-256 pod, for 15 and 4 hours in the first and second stages, respectively.
We align the dynamic point cloud sequence length with the robot's action chunk at length four, sample 200 points on the hand and gripper during dynamic point cloud training, and predict changes in point cloud positions. In simulation and on the real robot, we use end-effector control, consisting of delta end-effector cartesian positions, delta euler rotations, and a gripper closedness command. During inference, we use Euler integration with $\Delta t = 0.1$ to compute the predicted targets. When running the model on the WidowX robot, we execute the whole action chunk.

\subsection{Baselines}
We refer to our model, pre-trained on unlabeled data from BridgeData V2~\cite{BridgeV2}, Rt-1~\cite{brohan2022rt}, and RH20T~\cite{rh20t}, as \textbf{MotoVLA (R + H)} and compare against the following state-of-the-art methods.

\textbf{OpenVLA (OXE)~\cite{OpenVLA}} is a state-of-the-art open-source VLA and serves as our comparison to generalist VLAs trained on the full large-scale Open X-Embodiment dataset~\cite{oxe}. We only compare to OpenVLA on the real robot, since we find the model does not work well in the SIMPLER Bridge setting.

\textbf{LAPA (OXE)~\cite{LAPA}} also utilizes a two-stage training approach, but in a specialist policy setting. LAPA pretrains on latent actions from Open X-Embodiment~\cite{oxe}, and finetunes for specific tasks with robot actions. Since the model is only available for finetuning on specific tasks, we compare it to LAPA finetuned on 100 episodes, identical to the test setting, from the SIMPLER simulator. 

\textbf{$\pi_0$ (B)~\cite{pi0}} is a state-of-the-art generalist VLA with a similar architecture to our model. It serves as the baseline for validating the efficacy of the dynamic point cloud training for skill transfer. In order to facilitate a fair comparison, we train the $\pi_0$ model from scratch on BridgeData V2~\cite{BridgeV2}.

\textbf{ATM (B)~\cite{ATM}} is a hierarchical approach that first learns a 2D pixel track prediction model from video and then computes actions with a track conditioned policy. Since ATM is intended for small-scale imitation learning, we adapt the model for the zero-shot setting by replacing the track transformer with Paligemma and using flow-matching for track and action generation. We denote this modified ATM model as ATM (B). More details can be found in the Appendix~\ref{app:atm}.

\textbf{MotoVLA (R)} is our model trained without the human demonstration data in RH20T~\cite{rh20t}. It serves to quantify the effect of adding human demonstrations to the point cloud sequence pre-training stage.

\subsection{In-domain experiments}

\textbf{Experimental Setup.} This experiment evaluates the effect of dynamic point cloud pre-training on tasks and environments that are included in the action-labeled dataset. Thus, these tasks are in-domain for all models. In particular, we investigate whether the second training stage can retain and leverage the self-supervised representations learned during the first stage, or whether these representations are lost due to the direct action supervision. In order to facilitate reproducibility and fair comparison, we evaluate inside the SIMPLER~\cite{simpler} BridgeData V2~\cite{BridgeV2} simulation environment. The environment contains four tasks from the BridgeData V2 dataset~\cite{BridgeV2} that are recreated via visual matching and system identification of the WidowX 250S robot. We evaluate every task for 24 random object configurations.

\textbf{Results.} Results for this experimental setting are presented in Table~\ref{tab:SimplerEval}, our model MotoVLA (R + H) achieves the highest average score of $68.2\%$ in the SIMPLER simulator. It outperforms LAPA~\cite{LAPA} by $14.1\%$, even though LAPA is pre-trained on the whole Open X-Embodiment dataset and specifically finetuned with 100 episodes from the exact target task in simulation. Our method also outperforms $\pi_0$ (B) as a baseline model by $11.4\%$, showcasing that learning cross-domain and cross-embodiment motion priors via dynamic point cloud prediction translates to improved downstream performance even for tasks with action supervision in the second training stage. It also becomes evident from ATM (B), that using a hierarchical manipulation policy leads to significantly worse results in the generalist setting. 

Finally, we observe that MotoVLA (R + H) is marginally better than MotoVLA (R), mainly due to a large increase in cube stacking performance. Unlike the other tasks, cube stacking is well represented in the human demonstration data. 
However, to more explicitly study the effect of human data, we next examine the out-of-domain experiments.

\begin{table*}[tb]
    \centering
    \small
    \begin{tabular}{p{2.4cm} p{1.6cm} p{1.6cm} p{1.6cm} p{1.6cm} p{1.6cm}}
        \multirow{2}{*}{\makecell{Policy}} & \multicolumn{4}{c}{SIMPLER Tasks} \\
        \cmidrule(lr){2-5}
         &\shortstack[l]{Put carrot\\on plate}
         &\shortstack[l]{Put eggplant\\in basket}
         &\shortstack[l]{Put spoon\\on towel}
         &\shortstack[l]{Put gray on\\yellow block}
         &Average \\
        \midrule
        ATM (B) & $16.6\pm8.3$ & $43.8\pm6.2$ & $18.8\pm2.0$ & $4.1\pm4.1$ & $20.8\pm2.0$ \\
        LAPA (OXE) & $37.5\pm1.3$ & $50.0\pm2.7$ &  $70.8\pm4.1$ & $\mathbf{58.3\pm1.3}$ & $54.1 \pm2.3$ \\
        $\pi_0$ (B) & $39.6\pm6.2$ & $83.3\pm0.0$ & \underline{$72.9\pm6.2$} & $31.2\pm2.0$ & $56.8\pm0.5$ \\
        \addlinespace
        \midrule
        MotoVLA (R)  & $\mathbf{75.0\pm4.1}$ & $\mathbf{100.0\pm0.0}$ & $\mathbf{75.0\pm4.1}$ & $12.5\pm4.1$ & \underline{$65.6\pm3.1$} \\
        MotoVLA (R + H) & \underline{$54.1\pm1.3$} & \underline{$97.9\pm1.3$} & \underline{$72.9\pm3.6$} & \underline{$47.9\pm2.4$} & $\mathbf{68.2\pm0.9}$ \\
        \bottomrule
    \end{tabular}
    \caption{Mean success rate ($\pm$ standard error) of in-domain tasks in SIMPLER.}
    \label{tab:SimplerEval}
    \vspace{-5mm}
\end{table*}

\subsection{Out-of-domain experiments}
\textbf{Experiment Setup.} The out-of-domain setting aims to validate our model on tasks and in environments that the model has not seen during the action training stage. We differentiate between completely unseen tasks, which are neither contained in the dynamic point cloud pre-training nor the action training stage, and tasks that are out-of-domain for the action dataset, but in-domain for the unlabeled stage 1 dynamic point cloud training. Therefore, the first scenario tests for an overall improvement in generalization, while the second explicitly tests for an unsupervised skill transfer from unlabeled data to robot actions. In both cases, the tasks require visual generalization (to unseen backgrounds) and physical generalization (to unseen target object sizes and shapes), since it's not possible to exactly recreate the same environment and target objects as in the training data.

We evaluate each method for eight tasks with ten environment configurations and three trials each. More information about each task can be found in the Appendix~\ref{app:tasks}. We again use the WidowX 250S robot as an embodiment, but perform experiments on the real robot.

\begin{figure}[ht]
    \centering
    \includegraphics[width=0.9\columnwidth ,trim={8mm 4mm 8mm 3mm}, clip]{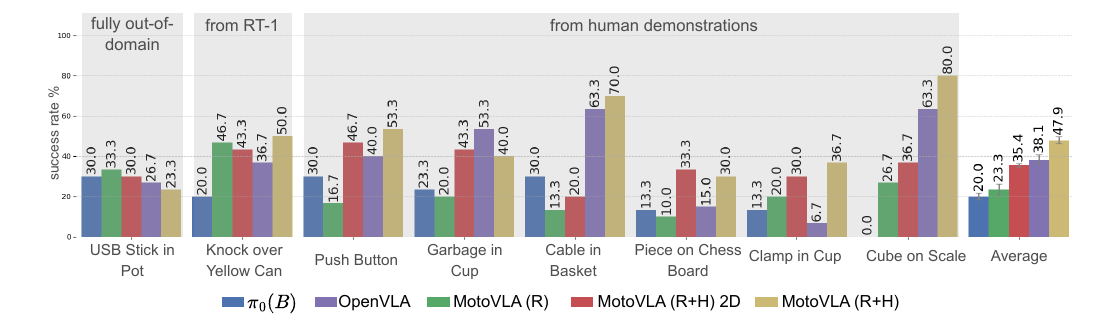}
    \caption{Complete overview of our results during the out-of-action domain evaluation on the real WidowX robot. We differentiate between completely out-of-domain tasks and tasks that are part of the Rt-1 actionless dataset or the RH20T human demonstration dataset. We report the mean success rate for all tasks and additionally the standard error for the average performance.}
    \label{fig:ood_performance}
\end{figure}

\textbf{Results.} We summarize the results in Figure~\ref{fig:ood_performance}. Our MotoVLA (R + H) model achieves the best average success rate over all out-of-domain tasks. Notably, the biggest gains in performance come from the tasks Push Button, Cube on Scale, Cable in Basket, and Clamp in Cup, which are present in the human demonstration dataset during dynamic point cloud training. This demonstrates that our approach can directly transfer knowledge on unseen objects and tasks from unlabeled data across embodiments.

In these tasks, we furthermore observe that the model pre-trained on corresponding human demonstrations exhibits faster and more precise trajectories, compared to MotoVLA (R) and the $\pi_0$-B baselines, which have never encountered these tasks. In these settings, the $\pi_0$-B baseline and MotoVLA (R) models have a higher tendency to encounter stuck states or move in an erratic direction. OpenVLA’s 9.8\% inferior performance, even with more robot data, highlights the value of task-specific human demonstrations, which are more scalable to collect than equivalent robot data.

Finally, for tasks that are completely out-of-domain, during stage 1 as well as stage 2 straining, the MotoVLA (R) model achieves the best results, with the $\pi_0$-B baseline following closely, showing that such fully out-of-domain tasks remain challenging. Overall, these results strongly support that including tasks as unlabeled human or robot demonstrations in the first training stage, improves downstream performance of VLAs on similar tasks, even if they are not included in the action-labeled dataset of the second stage training.  

\begin{wraptable}[7]{r}{0.5\columnwidth}
    \vspace{-30pt}
    \centering
    \footnotesize
    \begin{tabular}{p{4cm} p{1.8cm}}
        Method
        &Average \\
        \midrule
        MotoVLA Separation & 46.9  \\
        MotoVLA Co-Training  & 47.9 \\
        MotoVLA Grid  & 54.7 \\
        MotoVLA 32 Points & 60.9  \\
        MotoVLA (R + H) 2D  & 64.2  \\
        MotoVLA (R + H)  & \textbf{68.2} \\
    \end{tabular}
    \caption{Ablation study depicting the success rate of robotics foundation models in SIMPLER.}
    \label{tab:ablation}
\end{wraptable}
\vspace{-3mm}
\subsection{Ablations}
We evaluate the different design choices made during the development of our method by ablating the model on the same data mixture in the SIMPLER in-domain test setting. For the 2D ablation, we additionally evaluate in the real-world out-of-domain experiment. A full comparison of our ablation results in SIMPLER is depicted in Table~\ref{tab:ablation}, where we test the following:

\textbf{Two Dimensional Tracks} are evaluated with first-stage training to predict two-dimensional pixel positions instead of the three-dimensional dynamic point clouds. Using 2D tracks reduces performance by 4.0\% in SIMPLER and 12.5\% when evaluated on the real robot (see Figure~\ref{fig:ood_performance}). These results show the benefit of using 3D point clouds, which have a smaller domain gap when compared to robot end-effector actions. \\
\textbf{Track Points Selection} is evaluated by 1) using 32 instead of 200 points, and 2) sampling uniformly instead of on the gripper, which both show a strong performance drop.\\
\textbf{Predictor Co-Training} with dynamic point cloud prediction during the second stage is evaluated by training a predictor that jointly performs both tasks. While this approach indeed helps preserve the first-stage track features, it comes at the cost of degraded action prediction performance.\\
\textbf{Predictor Separation} is an alternative where two separate models for dynamic point cloud and action prediction are trained. While this approach is a natural modularization, it is ineffective in gaining knowledge from the unlabeled data. More architectural details are available in the Appendix~\ref{app:track_expert}.

\section{Conclusion}
In this work, we approached the challenge of learning a generalist manipulation policy from a mix of labeled and unlabeled video and proposed MotoVLA together with a two-stage training procedure. By establishing dynamic point clouds as an embodiment agnostic representation, our approach successfully transfers knowledge from video to manipulation motion priors. Using simulation and real-world experiments, we demonstrate a consistently improved model performance in in- and out-of-domain settings and showcase the direct transfer from human demonstration to robot actions.


\newpage

\section{Limitations}
\label{sec:limitations}
In this work, we presented MotoVLA, a generalist manipulation policy trained on both labeled and unlabeled human and robot demonstrations. We showed that training for point cloud sequence prediction on this unlabeled data mixture allows the model to generalize to tasks unseen during the action training phase. 

However, the current approach also comes with its limitations. First, the human demonstrations in RH20T are fairly close to the robotics domain, with a static third-person camera and background. Most open-world human demonstration datasets~\cite{EpicKitchen, Ego4D} are egocentric with non-static head-mounted cameras. In order to extract the point cloud sequence with respect to a static camera frame, one needs to additionally estimate the camera extrinsics using methods like MonST3R~\cite{Monster}. Another problem occurs when hand movement during human demonstrations is very rapid, causing the TAP model to lose track of its query points. With ever-improving TAP models, this issue will likely be reduced. Nevertheless, careful data curation will remain important for maximizing downstream performance. Moreover, while current monocular depth estimators provide impressive results, achieving consistency and geometric accuracy, especially for diverse backgrounds, is still challenging. Future improvements in these areas would likely also yield better results with our method. Overall, while we show significant performance gains over baseline methods for both in-domain and out-of-domain settings, the further improvement also relies on the improvements in the domains beyond the scope of this paper, such as egocentric video understanding and recording of fast motions with ease. We hope that our work will enable the community to make further progress towards achieving robust and generalizable robot manipulation. 

\acknowledgments{ This research was partially funded by the Ministry of Education and Science of Bulgaria (support for INSAIT, part of the Bulgarian National Roadmap for Research Infrastructure). This project was also supported with computational resources provided by Google Cloud Platform (GCP).}

\bibliography{example}  
\appendix
\input{appendix}
\end{document}

%% file: appendix.tex
\section{Training Data}
\label{app:training_data}
\textbf{Training Data Mixture.}
We use the following datasets and mixture weights during Dynamic Point Cloud Training:

\begin{table}[h]
    \centering
    \footnotesize
    \begin{tabular}{p{4cm} p{1.8cm}}
        \toprule
        Dataset & Weight \\
        \midrule
        Rt-1~\cite{brohan2022rt} & 1.0  \\
        BridgeData V2~\cite{BridgeV2}  & 1.0 \\
        RH20T (Human)~\cite{rh20t} & 1.0 \\
        \bottomrule
    \end{tabular}
    \vspace{2mm}
    \caption{The datasets used during Dynamic Point Cloud Training and their corresponding mixture weight.}
    \label{tab:dataset_weights}
\end{table}
\vspace{-10pt}
We filter out episodes with missing language annotation or an insufficient number of unique points in the point cloud. During Action Alignment, we only train on actions from BridgeData V2~\cite{BridgeV2}.

\section{Training Details}
\textbf{Dynamic Point Cloud Training} uses the following hyperparameters:

\begin{table}[h]
    \centering
    \footnotesize
    \begin{tabular}{p{4cm} p{4cm}}
        \toprule
 Hyperparameteres & 3D Dynamics Predictor \\
        \midrule
 epochs & 20  \\
 batch size  & 2048 \\
 optimizer & Adam \\
 learning rate & 5e-5 \\
 weight decay & 0.0 \\
 lr scheduler & const \\
 lr warmup steps & 16000 \\
 clip grad & 1.0 \\
 number of points & 200 \\
 point cloud sequence length & 4 \\
 image augmentations & ColorJitter, RandomCrop \\
        \bottomrule
    \end{tabular}
    \vspace{2mm}
    \caption{The training hyperparameters used during Dynamic Point Cloud Training.}
    \label{tab:dynamic_pointcloud_training_params}
\end{table}

\textbf{Action Alignment} training uses the following differing hyperparameters:

\begin{table}[h]
    \centering
    \footnotesize
    \begin{tabular}{p{4cm} p{4cm}}
        \toprule
 Hyperparameteres & Action Predictor \\
        \midrule
 batch size  & 1024 \\
 lr warmup steps & 200 \\
 action chunk length & 4.0 \\
        \bottomrule
    \end{tabular}
    \vspace{2mm}
    \caption{The training hyperparameters used during Action Alignment.}
    \label{tab:action_align_training_params}
\end{table}

\section{Architecture Details}
\label{app:architecture}
Our 3D Dynamics Predictor and Action Predictor are based on the Gemma 2B language model architecture, with several modifications to reduce the overall parameter count. Specifically, we reduce the feature dimensionality to 1024 and set the MLP dimensionality within the attention blocks to 4096. To encode the flow-matching timestep, we use sinusoidal embeddings, which are concatenated with the noise embeddings. This combined representation is then projected to the Transformer feature dimension through two linear layers, separated by a Swish activation function. We adopt the following attention pattern: Paligemma tokens use bidirectional attention among themselves but cannot attend to the tokens from the 3D Dynamics Predictor or the Action Predictor. In contrast, the 3D Dynamics Predictor and Action Predictor tokens have full attention over all other tokens, with one exception—tokens corresponding to the current point cloud and proprioception are restricted from attending to the noise tokens.

\section{Ablations}

\subsection{Predictor Co-Training}
\label{app:track_co}

\begin{figure}
    \centering
    \includegraphics[width=0.5\columnwidth]{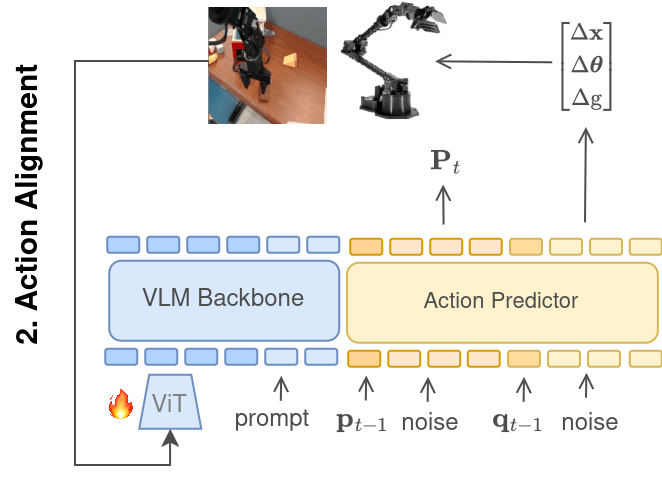}
    \caption{The architecture used during our Predictor Co-Training experiment. The Action Predictor predicts the dynamic point cloud $\mathbf{P}_t$ and robot actions in parallel.}
    \label{fig:track_co}
\end{figure}

During Predictor Co-Training, we train the 3D Dynamics predictor in the same manner as introduced in Section~\ref{subsec:architecture}, but modify the action alignment training stage and inference procedure. During action alignment, we predict both the dynamic point cloud and the robot action denoising vector field in parallel. We do this, by conditioning on both the current point cloud and current proprioception, as well as the respective noisy samples. The linear encoders and decoders of the proprioception and action noise get initialized from scratch, while the ones for the dynamic point clouds are reused from the dynamic point cloud training stage. The training loss is the combination of the previously introduced losses: $\mathcal{L}_{tot} = \mathcal{L}_{points} + \mathcal{L}_{act}$. 
During inference, we estimate the gripper point cloud in real-time by first computing a segmentation mask of the gripper using Grounding Dino~\cite{GroundingDINO} and Segment Anything 2~\cite{SAM2}, sampling 200 points on the mask, and lifting them to 3D using MoGE~\cite{MoGE}. We then perform flow-matching~\cite{FlowMatching} inference by Euler integrating both the dynamic point cloud and action denoising vectors with $\Delta t=0.1$. An overview of the predictor co-training action alignment architecture can be found in Figure~\ref{fig:track_co}.

\subsection{Predictor Separation}
\label{app:track_expert}
\begin{figure}[H]
    \centering
    \includegraphics[width=0.5\columnwidth, clip]{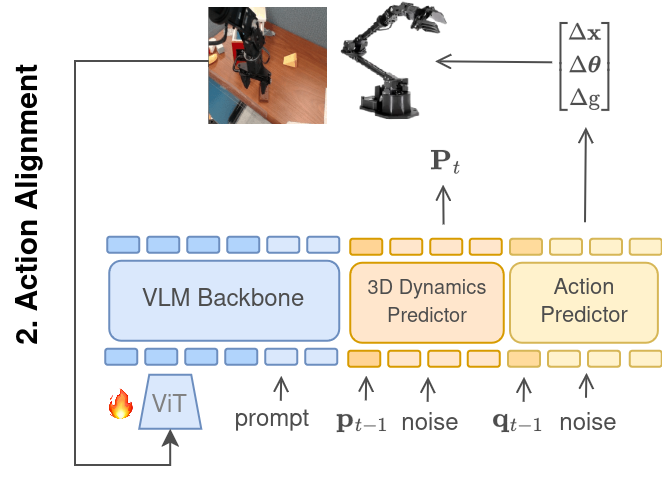}
    \caption{The architecture used during our Predictor Expert Separation ablation experiment. The 3D Dynamics- and Action Predictor predict the dynamic point cloud $\mathbf{P}_t$ and robot actions in parallel.}
    \label{fig:track_expert}
\end{figure}

During the Predictor Separation, we do not modify the dynamic point cloud training stage, but make the following modifications in the action alignment stage. Instead of initializing the Action Predictor from the 3D Dynamics Predictor, we introduce a randomly initialized Action Predictor while keeping the 3D Dynamics Predictor unchanged.  The resulting model is a Mixture of Transformers~\cite{MixtureOfTransformers} with three separate components, the VLM, the 3D Dynamics Predictor, and the Action Predictor. The Action Predictor attends to the 3D Dynamics Predictor and the Vision-Language Model (VLM) key-value cache. Additionally, the 3D Dynamics Predictor is conditioned on the current point cloud and noisy future point cloud sample, while the Action Predictor is conditioned on the proprioception and noisy action sample. During flow-matching, the 3D Dynamics Predictor and Action Predictor predict the denoising vector for their respective features in parallel. The total training loss is computed as: $\mathcal{L}_{tot} = \mathcal{L}_{points} + \mathcal{L}_{act}$. During inference, the current point cloud is computed in real-time by using the same approach as introduced in Subsection~\ref{app:track_co} 
Figure~\ref{fig:track_expert} provides an overview of the architecture.

\section{Experimental Results}

\subsection{Baselines}
\subsubsection{ATM (B)}
\label{app:atm}
The original ATM~\cite{ATM} model utilizes a bidirectional track transformer to predict 2D pixel tracks from a padded sequence of current 2D pixel positions during its initial training phase. In the second phase, the Track Transformer is frozen, and a policy network is trained conditioned on the predicted tracks. Tailored for small-scale imitation learning, the original ATM model comprises a 20-million-parameter Track Transformer and vision encoder, trained from scratch during the first stage.

\begin{figure}[H]
    \centering
    \includegraphics[width=0.5\columnwidth, clip]{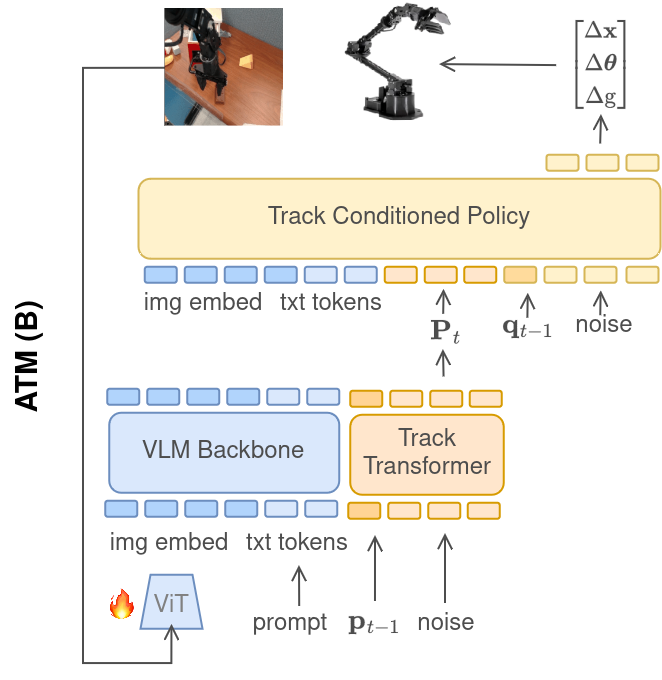}
    \caption{The architecture of our ATM (B) Baseline. First, the Track Transformer predicts 2D pixel tracks and then the track conditioned policy predicts robot actions. }
    \label{fig:atm_b}
\end{figure}

We introduce several key modifications to adapt this approach to the zero-shot setting and establish a fair baseline. First, we condition the track transformer on the key-value cache of a pre-trained Paligemma model via a Mixture of Transformers architecture~\cite{MixtureOfTransformers}. Additionally, we use flow-matching to generate tracks and actions in the track transformer and track-conditioned policy respectively. In the second training stage, we freeze the Paligemma VLM and track transformer and run full flow-matching inference of the track transformer with Euler integration and $\Delta t = 0.1$. We then condition the policy on the generated 2D-pixel tracks, text embeddings, and SigLip features. The track transformer and track-conditioned policy are the same downscaled Gemma~\cite{gemma} architecture used for the 3D Dynamics Predictor and Action Predictor. We encode the 2D-pixel tracks via a linear layer and use the same data mixture throughout both training stages as for our MotoVLA (R + H) model. A full overview of the ATM (B) architecture can be found in Figure~\ref{fig:atm_b}.

\subsection{Out-of-domain experiments}
\subsubsection{Tasks}
\label{app:tasks}

\begin{figure}
    \centering
    \includegraphics[width=0.4\columnwidth, clip]{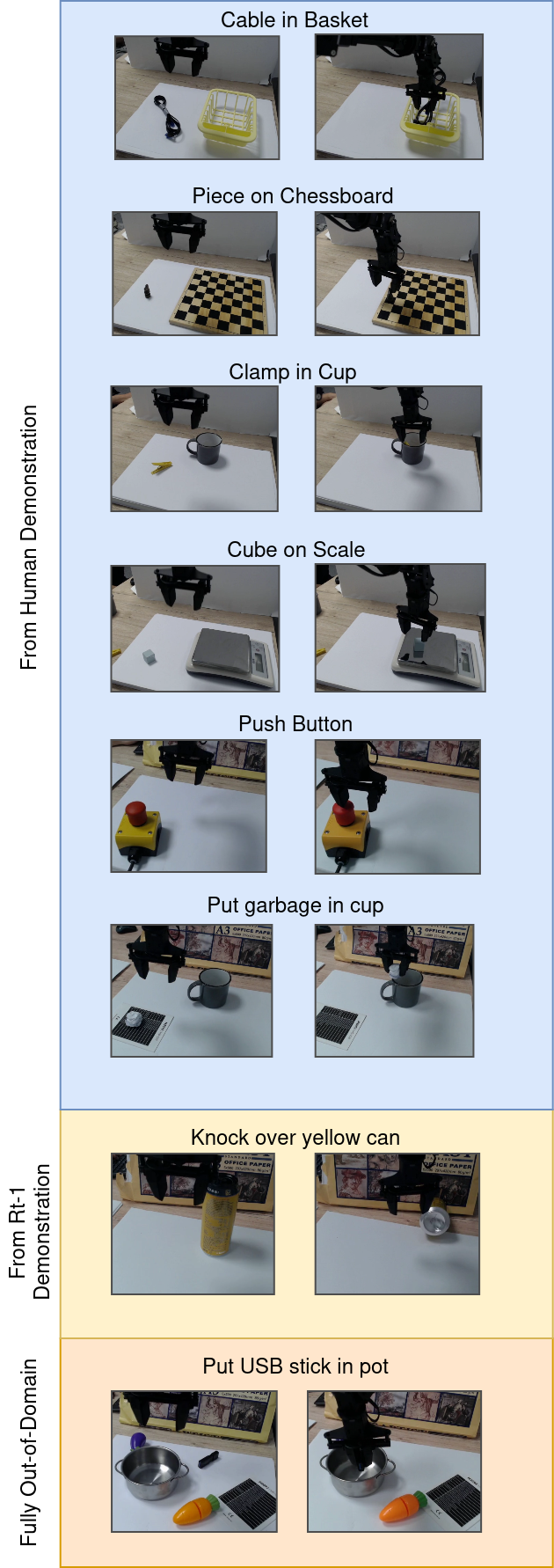}
    \caption{The different tasks with the start, end frame, and category.}
    \label{fig:task_overview}
\end{figure}

We roll out all our tasks in the same environment, with a white work surface and background containing white or colored images. None of the exact target objects were part of the training mixture. Accordingly, all tasks need some level of \textbf{visual} (unseen backgrounds, distractor objects, colors/appearances of objects) generalization. Figure~\ref{fig:task_overview} shows an overview of the tasks and their environments.
\\    

\begin{enumerate}
    \item \textbf{Put USB Stick in Pot.}
    This task aims to pick up a USB stick and place it in the pot. Additionally, there are some distractor objects from the BridgeData V2~\cite{BridgeV2} dataset, namely an eggplant and a carrot. Since this task is not part of any pre-training dataset, we consider it fully out-of-domain. It requires \textbf{physical} (unseen object sizes/shapes), \textbf{motion} (unseen object positions/orientations), and \textbf{semantic} (unseen target objects, instructions, and concepts from the Internet) generalization,  as well as robustness towards overfitting to the BridgeData V2~\cite{BridgeV2} objects.

    \item \textbf{Knock over Yellow Can.} This task aims to move the gripper to the yellow can and knock it over. The Rt-1 dataset~\cite{brohan2022rt} contains very similar tasks, with different types of cans. Accordingly, the task tests \textbf{physical}, \textbf{motion}, and \textbf{semantic} transfer from actionless cross-embodiment robot demonstrations.
    
    \item \textbf{Push Button.} This task aims to position the gripper over a red button and press it. Very similar tasks exist in the RH20T~\cite{rh20t} dataset. Accordingly, the task tests \textbf{physical}, \textbf{motion}, and \textbf{semantic} transfer from actionless cross-embodiment human demonstrations.

    \item \textbf{Put Garbage in Cup.} The goal here is to grasp a crumpled piece of paper and place it in the cup. Again, similar tasks exist in the RH20T~\cite{rh20t} dataset. Accordingly, the task tests \textbf{physical}, \textbf{motion} and \textbf{semantic} transfer from actionless cross-embodiment

    \item \textbf{Cable in Basket.} In this task the robot needs to grasp a cable roll and place it in a basket. RH20T~\cite{rh20t} contains similar cable manipulation tasks, however, our cable differs in shape and size from the dataset. Accordingly, the task tests \textbf{physical} generalization and both \textbf{motion} and $\textbf{semantic}$ transfer from the unlabeled human demonstrations.

    \item \textbf{Piece on Chessboard.} Here, the robot needs to place a chess piece on the empty chessboard next to it. Again we adapt this task from RH20T~\cite{rh20t}. We always use the same chess piece and give partial success (0.5), if the model grasps the piece, and full success (1.0), if the model releases it on the chessboard without it falling over. The task tests \textbf{physical}, $\textbf{motion}$, and $\textbf{semantic}$ transfer from unlabeled human demonstrations. 

    \item \textbf{Clamp in Cup.} This task, again from RH20T~\cite{rh20t}, demands the robot to grasp a laundry clamp and put it in a cup. It tests \textbf{physical} and $\textbf{motion}$ transfer from the unlabeled demonstrations.

    \item \textbf{Cube on Scale.} Again adapted from RH20T~\cite{rh20t}, the goal of this task is to pick up a cube and place it on the scale next to it. The task again tests \textbf{physical}, $\textbf{motion}$, and $\textbf{semantic}$ transfer from unlabeled human demonstrations.

\end{enumerate}

\clearpage

\subsubsection{Rollouts:}
\label{app:full_roll}
In the following Figure \ref{fig:ood_rollouts}, we present image sequences from full rollouts of our MotoVLA (R+H) model and the $\pi_0 (B)$ baseline in all eight real-world out-of-action-domain tasks. 

\begin{figure}[H]
    \centering
    \includegraphics[width=0.85\columnwidth]{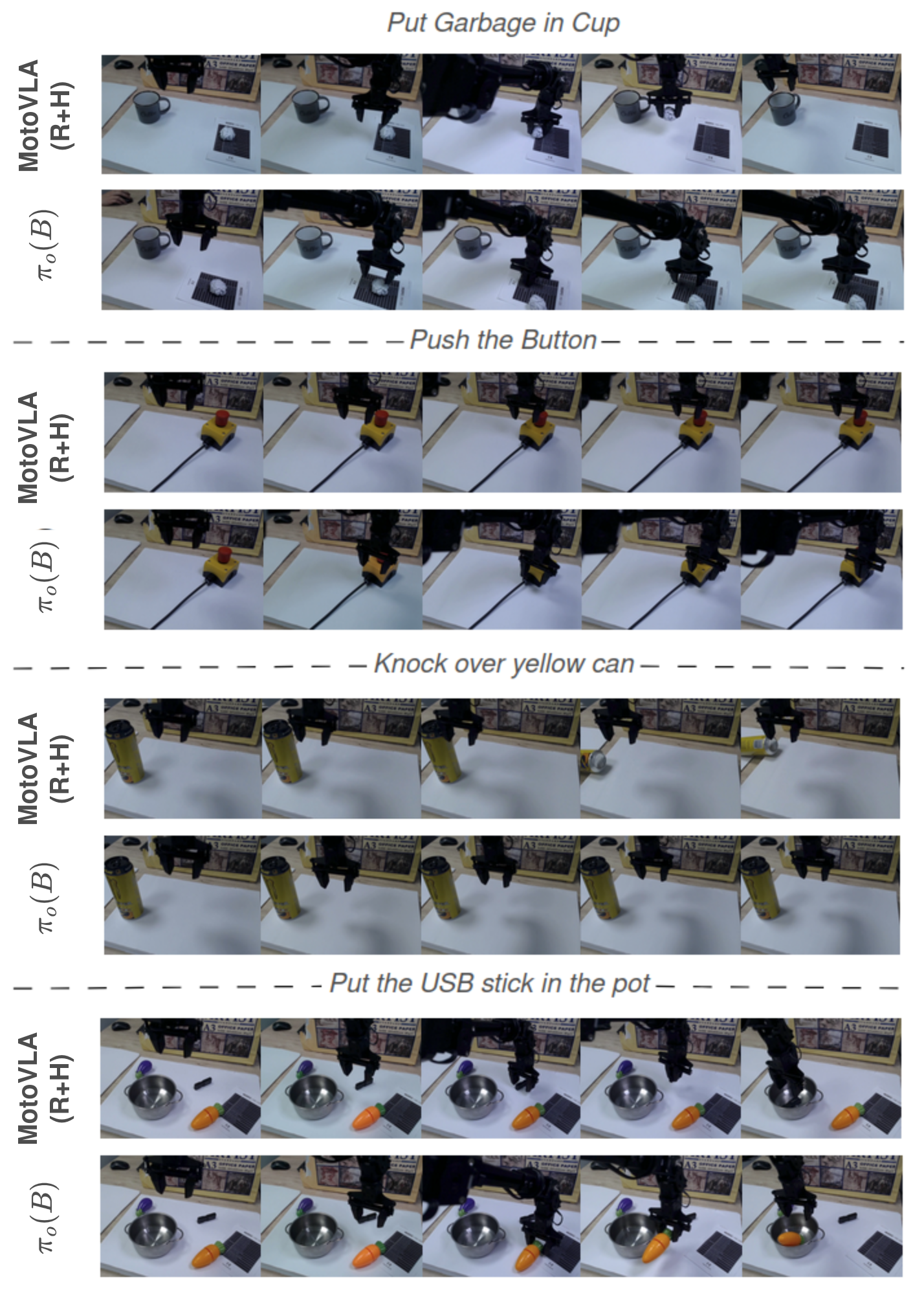}
    \label{fig:ood_rollouts1}
\end{figure}

\begin{figure}[H]
    \centering
    \includegraphics[width=0.9\columnwidth]{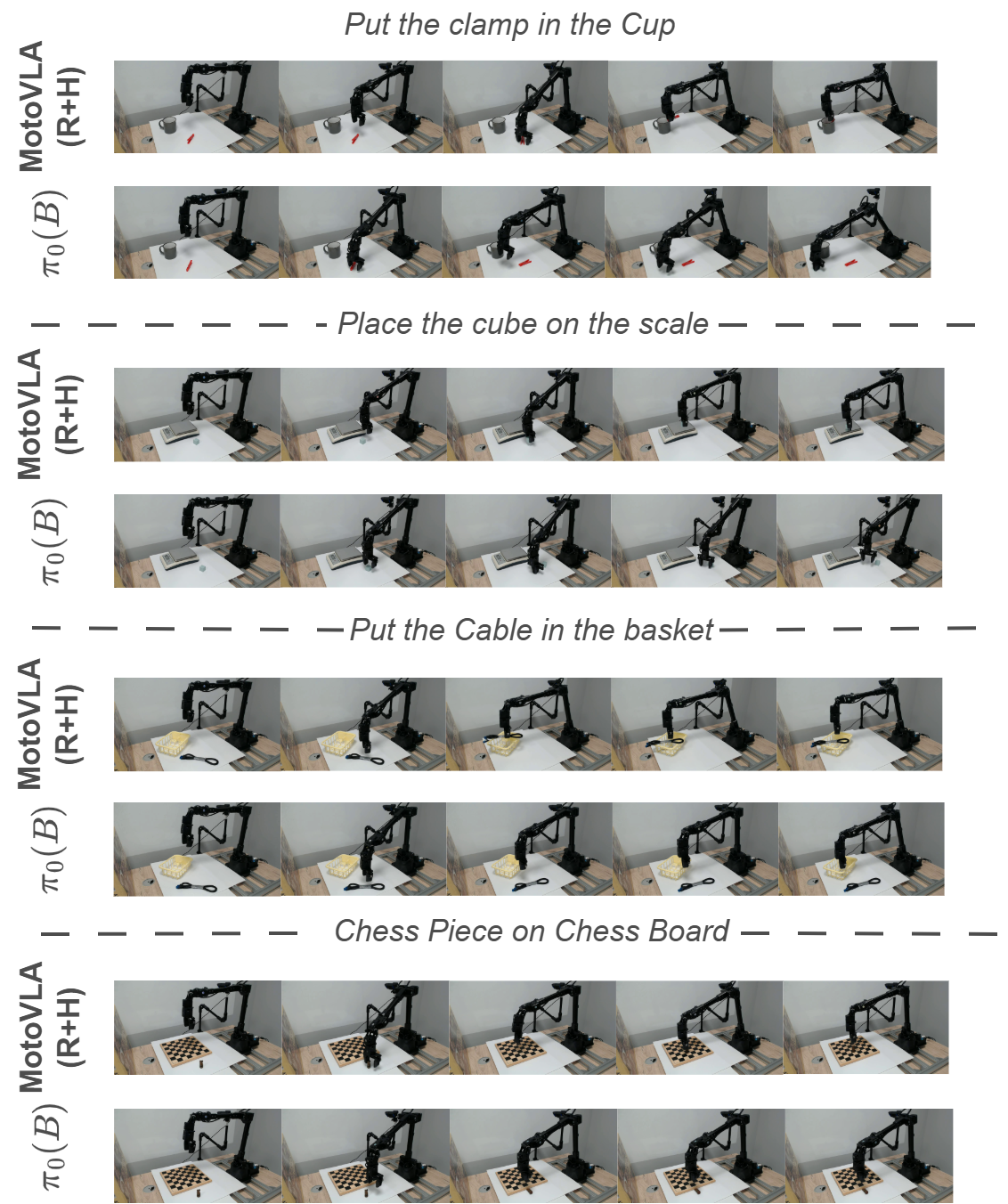}
    \caption{The image sequences show full rollouts of our MotoVLA (R+H) model and the $\pi_0 (B)$ baseline for eight of the out-of-domain tasks. The $\pi_0 (B)$ baseline struggles with grasping precision and displays more incoherent behavior.}
    \label{fig:ood_rollouts}
\end{figure}

\subsection{In-domain Rollouts}
\label{app:ind}

In the following Figure \ref{fig:id_rollouts}, we show image sequences from full rollouts of the four BridgeData V2~\cite{BridgeV2} tasks inside the SIMPLER~\cite{simpler} simulator.

\begin{figure}
    \centering
    \includegraphics[width=0.7\columnwidth]{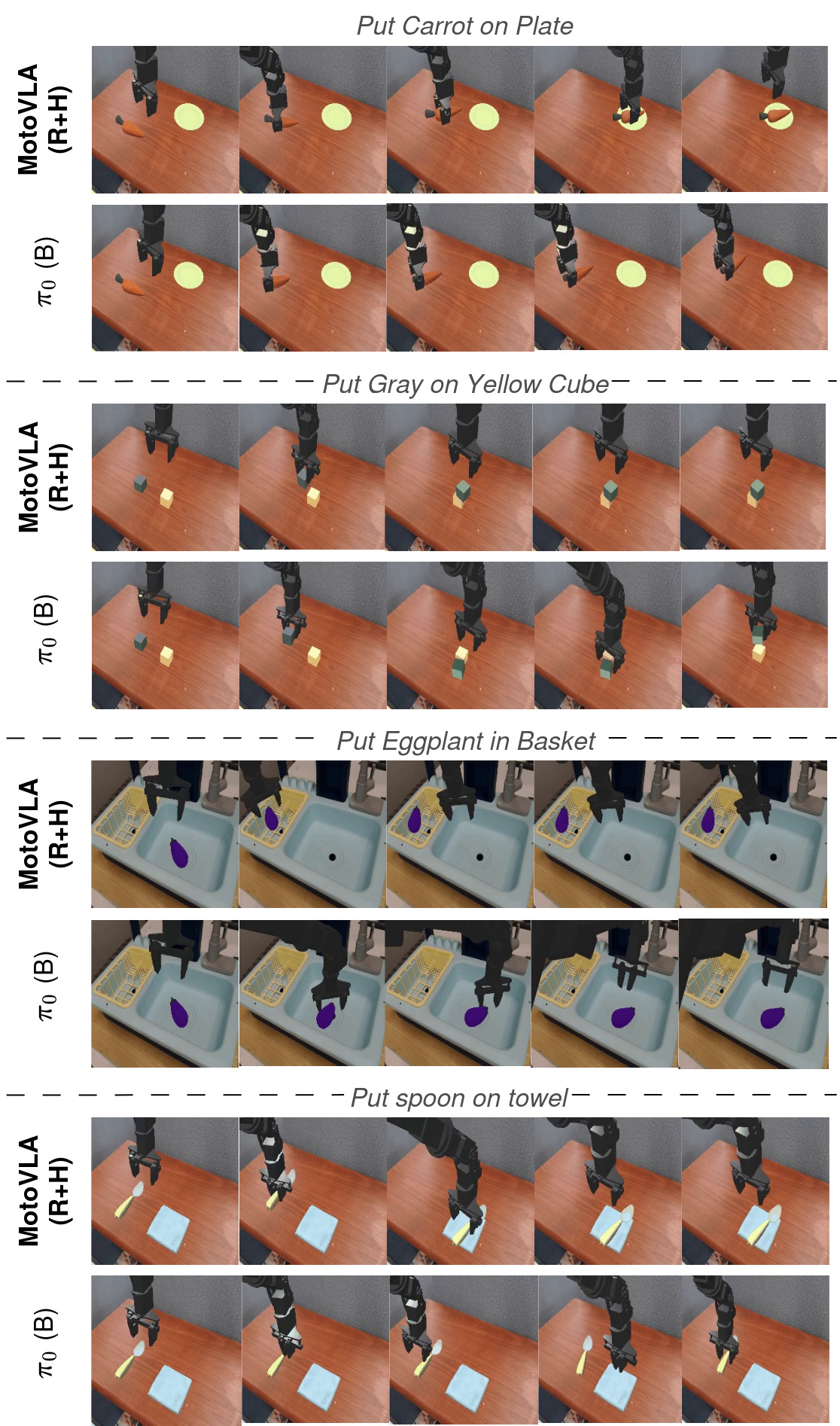}
    \caption{The image sequences show full rollouts of our MotoVLA (R+H) model and the $\pi_0 (B)$ baseline for out-of-domain tasks. The $\pi_0 (B)$ baseline has worse grasping precision.}
    \label{fig:id_rollouts}
\end{figure}

%% file: arxiv.bbl
\begin{thebibliography}{53}
\providecommand{\natexlab}[1]{#1}
\providecommand{\url}[1]{\texttt{#1}}
\expandafter\ifx\csname urlstyle\endcsname\relax
  \providecommand{\doi}[1]{doi: #1}\else
  \providecommand{\doi}{doi: \begingroup \urlstyle{rm}\Url}\fi

\bibitem[Kim et~al.(2025)Kim, Pertsch, Karamcheti, Xiao, Balakrishna, Nair, Rafailov, Foster, Sanketi, Vuong, Kollar, Burchfiel, Tedrake, Sadigh, Levine, Liang, and Finn]{OpenVLA}
M.~J. Kim, K.~Pertsch, S.~Karamcheti, T.~Xiao, A.~Balakrishna, S.~Nair, R.~Rafailov, E.~P. Foster, P.~R. Sanketi, Q.~Vuong, T.~Kollar, B.~Burchfiel, R.~Tedrake, D.~Sadigh, S.~Levine, P.~Liang, and C.~Finn.
\newblock Openvla: An open-source vision-language-action model.
\newblock In P.~Agrawal, O.~Kroemer, and W.~Burgard, editors, \emph{Proceedings of The 8th Conference on Robot Learning}, volume 270 of \emph{Proceedings of Machine Learning Research}, pages 2679--2713. PMLR, 06--09 Nov 2025.
\newblock URL \url{https://proceedings.mlr.press/v270/kim25c.html}.

\bibitem[Black et~al.(2024)Black, Brown, Driess, Esmail, Equi, Finn, Fusai, Groom, Hausman, Ichter, Jakubczak, Jones, Ke, Levine, Li-Bell, Mothukuri, Nair, Pertsch, Shi, Tanner, Vuong, Walling, Wang, and Zhilinsky]{pi0}
K.~Black, N.~Brown, D.~Driess, A.~Esmail, M.~Equi, C.~Finn, N.~Fusai, L.~Groom, K.~Hausman, B.~Ichter, S.~Jakubczak, T.~Jones, L.~Ke, S.~Levine, A.~Li-Bell, M.~Mothukuri, S.~Nair, K.~Pertsch, L.~X. Shi, J.~Tanner, Q.~Vuong, A.~Walling, H.~Wang, and U.~Zhilinsky.
\newblock $\pi_0$: A vision-language-action flow model for general robot control, 2024.
\newblock URL \url{https://arxiv.org/abs/2410.24164}.

\bibitem[Collaboration et~al.(2023)Collaboration, O'Neill, Rehman, Gupta, Maddukuri, Gupta, Padalkar, Lee, Pooley, Gupta, Mandlekar, Jain, Tung, Bewley, Herzog, Irpan, Khazatsky, Rai, Gupta, Wang, Kolobov, Singh, Garg, Kembhavi, Xie, Brohan, Raffin, Sharma, Yavary, Jain, Balakrishna, Wahid, Burgess-Limerick, Kim, Schölkopf, Wulfe, Ichter, Lu, Xu, Le, Finn, Wang, Xu, Chi, Huang, Chan, Agia, Pan, Fu, Devin, Xu, Morton, Driess, Chen, Pathak, Shah, Büchler, Jayaraman, Kalashnikov, Sadigh, Johns, Foster, Liu, Ceola, Xia, Zhao, Frujeri, Stulp, Zhou, Sukhatme, Salhotra, Yan, Feng, Schiavi, Berseth, Kahn, Yang, Wang, Su, Fang, Shi, Bao, Amor, Christensen, Furuta, Bharadhwaj, Walke, Fang, Ha, Mordatch, Radosavovic, Leal, Liang, Abou-Chakra, Kim, Drake, Peters, Schneider, Hsu, Vakil, Bohg, Bingham, Wu, Gao, Hu, Wu, Wu, Sun, Luo, Gu, Tan, Oh, Wu, Lu, Yang, Malik, Silvério, Hejna, Booher, Tompson, Yang, Salvador, Lim, Han, Wang, Rao, Pertsch, Hausman, Go, Gopalakrishnan, Goldberg, Byrne, Oslund, Kawaharazuka, Black, Lin, Zhang, Ehsani, Lekkala, Ellis, Rana, Srinivasan, Fang, Singh, Zeng, Hatch, Hsu, Itti, Chen, Pinto, Fei-Fei, Tan, Fan, Ott, Lee, Weihs, Chen, Lepert, Memmel, Tomizuka, Itkina, Castro, Spero, Du, Ahn, Yip, Zhang, Ding, Heo, Srirama, Sharma, Kim, Kanazawa, Hansen, Heess, Joshi, Suenderhauf, Liu, Palo, Shafiullah, Mees, Kroemer, Bastani, Sanketi, Miller, Yin, Wohlhart, Xu, Fagan, Mitrano, Sermanet, Abbeel, Sundaresan, Chen, Vuong, Rafailov, Tian, Doshi, Mart{'i}n-Mart{'i}n, Baijal, Scalise, Hendrix, Lin, Qian, Zhang, Mendonca, Shah, Hoque, Julian, Bustamante, Kirmani, Levine, Lin, Moore, Bahl, Dass, Sonawani, Tulsiani, Song, Xu, Haldar, Karamcheti, Adebola, Guist, Nasiriany, Schaal, Welker, Tian, Ramamoorthy, Dasari, Belkhale, Park, Nair, Mirchandani, Osa, Gupta, Harada, Matsushima, Xiao, Kollar, Yu, Ding, Davchev, Zhao, Armstrong, Darrell, Chung, Jain, Kumar, Vanhoucke, Zhan, Zhou, Burgard, Chen, Chen, Wang, Zhu, Geng, Liu, Liangwei, Li, Pang, Lu, Ma, Kim, Chebotar, Zhou, Zhu, Wu, Xu, Wang, Bisk, Dou, Cho, Lee, Cui, Cao, Wu, Tang, Zhu, Zhang, Jiang, Li, Li, Iwasawa, Matsuo, Ma, Xu, Cui, Zhang, Fu, and Lin]{oxe}
O.~X.-E. Collaboration, A.~O'Neill, A.~Rehman, A.~Gupta, A.~Maddukuri, A.~Gupta, A.~Padalkar, A.~Lee, A.~Pooley, A.~Gupta, A.~Mandlekar, A.~Jain, A.~Tung, A.~Bewley, A.~Herzog, A.~Irpan, A.~Khazatsky, A.~Rai, A.~Gupta, A.~Wang, A.~Kolobov, A.~Singh, A.~Garg, A.~Kembhavi, A.~Xie, A.~Brohan, A.~Raffin, A.~Sharma, A.~Yavary, A.~Jain, A.~Balakrishna, A.~Wahid, B.~Burgess-Limerick, B.~Kim, B.~Schölkopf, B.~Wulfe, B.~Ichter, C.~Lu, C.~Xu, C.~Le, C.~Finn, C.~Wang, C.~Xu, C.~Chi, C.~Huang, C.~Chan, C.~Agia, C.~Pan, C.~Fu, C.~Devin, D.~Xu, D.~Morton, D.~Driess, D.~Chen, D.~Pathak, D.~Shah, D.~Büchler, D.~Jayaraman, D.~Kalashnikov, D.~Sadigh, E.~Johns, E.~Foster, F.~Liu, F.~Ceola, F.~Xia, F.~Zhao, F.~V. Frujeri, F.~Stulp, G.~Zhou, G.~S. Sukhatme, G.~Salhotra, G.~Yan, G.~Feng, G.~Schiavi, G.~Berseth, G.~Kahn, G.~Yang, G.~Wang, H.~Su, H.-S. Fang, H.~Shi, H.~Bao, H.~B. Amor, H.~I. Christensen, H.~Furuta, H.~Bharadhwaj, H.~Walke, H.~Fang, H.~Ha, I.~Mordatch, I.~Radosavovic, I.~Leal, J.~Liang, J.~Abou-Chakra, J.~Kim, J.~Drake, J.~Peters, J.~Schneider, J.~Hsu, J.~Vakil, J.~Bohg, J.~Bingham, J.~Wu, J.~Gao, J.~Hu, J.~Wu, J.~Wu, J.~Sun, J.~Luo, J.~Gu, J.~Tan, J.~Oh, J.~Wu, J.~Lu, J.~Yang, J.~Malik, J.~Silvério, J.~Hejna, J.~Booher, J.~Tompson, J.~Yang, J.~Salvador, J.~J. Lim, J.~Han, K.~Wang, K.~Rao, K.~Pertsch, K.~Hausman, K.~Go, K.~Gopalakrishnan, K.~Goldberg, K.~Byrne, K.~Oslund, K.~Kawaharazuka, K.~Black, K.~Lin, K.~Zhang, K.~Ehsani, K.~Lekkala, K.~Ellis, K.~Rana, K.~Srinivasan, K.~Fang, K.~P. Singh, K.-H. Zeng, K.~Hatch, K.~Hsu, L.~Itti, L.~Y. Chen, L.~Pinto, L.~Fei-Fei, L.~Tan, L.~J. Fan, L.~Ott, L.~Lee, L.~Weihs, M.~Chen, M.~Lepert, M.~Memmel, M.~Tomizuka, M.~Itkina, M.~G. Castro, M.~Spero, M.~Du, M.~Ahn, M.~C. Yip, M.~Zhang, M.~Ding, M.~Heo, M.~K. Srirama, M.~Sharma, M.~J. Kim, N.~Kanazawa, N.~Hansen, N.~Heess, N.~J. Joshi, N.~Suenderhauf, N.~Liu, N.~D. Palo, N.~M.~M. Shafiullah, O.~Mees, O.~Kroemer, O.~Bastani, P.~R. Sanketi, P.~T. Miller, P.~Yin, P.~Wohlhart, P.~Xu, P.~D. Fagan, P.~Mitrano, P.~Sermanet, P.~Abbeel, P.~Sundaresan, Q.~Chen, Q.~Vuong, R.~Rafailov, R.~Tian, R.~Doshi, R.~Mart{'i}n-Mart{'i}n, R.~Baijal, R.~Scalise, R.~Hendrix, R.~Lin, R.~Qian, R.~Zhang, R.~Mendonca, R.~Shah, R.~Hoque, R.~Julian, S.~Bustamante, S.~Kirmani, S.~Levine, S.~Lin, S.~Moore, S.~Bahl, S.~Dass, S.~Sonawani, S.~Tulsiani, S.~Song, S.~Xu, S.~Haldar, S.~Karamcheti, S.~Adebola, S.~Guist, S.~Nasiriany, S.~Schaal, S.~Welker, S.~Tian, S.~Ramamoorthy, S.~Dasari, S.~Belkhale, S.~Park, S.~Nair, S.~Mirchandani, T.~Osa, T.~Gupta, T.~Harada, T.~Matsushima, T.~Xiao, T.~Kollar, T.~Yu, T.~Ding, T.~Davchev, T.~Z. Zhao, T.~Armstrong, T.~Darrell, T.~Chung, V.~Jain, V.~Kumar, V.~Vanhoucke, W.~Zhan, W.~Zhou, W.~Burgard, X.~Chen, X.~Chen, X.~Wang, X.~Zhu, X.~Geng, X.~Liu, X.~Liangwei, X.~Li, Y.~Pang, Y.~Lu, Y.~J. Ma, Y.~Kim, Y.~Chebotar, Y.~Zhou, Y.~Zhu, Y.~Wu, Y.~Xu, Y.~Wang, Y.~Bisk, Y.~Dou, Y.~Cho, Y.~Lee, Y.~Cui, Y.~Cao, Y.-H. Wu, Y.~Tang, Y.~Zhu, Y.~Zhang, Y.~Jiang, Y.~Li, Y.~Li, Y.~Iwasawa, Y.~Matsuo, Z.~Ma, Z.~Xu, Z.~J. Cui, Z.~Zhang, Z.~Fu, and Z.~Lin.
\newblock Open {X-E}mbodiment: Robotic learning datasets and {RT-X} models.
\newblock \url{https://arxiv.org/abs/2310.08864}, 2023.

\bibitem[Zhao et~al.(2025)Zhao, Lu, Kim, Fu, Zhang, Wu, Li, Ma, Han, Finn, Handa, Liu, Xiang, Wetzstein, and Lin]{cotvla}
Q.~Zhao, Y.~Lu, M.~J. Kim, Z.~Fu, Z.~Zhang, Y.~Wu, Z.~Li, Q.~Ma, S.~Han, C.~Finn, A.~Handa, M.-Y. Liu, D.~Xiang, G.~Wetzstein, and T.-Y. Lin.
\newblock Cot-vla: Visual chain-of-thought reasoning for vision-language-action models, 2025.
\newblock URL \url{https://arxiv.org/abs/2503.22020}.

\bibitem[Li et~al.(2024)Li, Liang, Wang, Luo, Chen, Liao, Wei, Deng, Xu, Zhang, Wang, Liu, Fu, Bao, Chen, Shi, Yang, and Guo]{cogact}
Q.~Li, Y.~Liang, Z.~Wang, L.~Luo, X.~Chen, M.~Liao, F.~Wei, Y.~Deng, S.~Xu, Y.~Zhang, X.~Wang, B.~Liu, J.~Fu, J.~Bao, D.~Chen, Y.~Shi, J.~Yang, and B.~Guo.
\newblock Cogact: A foundational vision-language-action model for synergizing cognition and action in robotic manipulation, 2024.
\newblock URL \url{https://arxiv.org/abs/2411.19650}.

\bibitem[Dey et~al.(2025)Dey, Zaech, Nikolov, Gool, and Paudel]{revla}
S.~Dey, J.-N. Zaech, N.~Nikolov, L.~V. Gool, and D.~P. Paudel.
\newblock Revla: Reverting visual domain limitation of robotic foundation models, 2025.
\newblock URL \url{https://arxiv.org/abs/2409.15250}.

\bibitem[Qu et~al.(2025)Qu, Song, Chen, Yao, Ye, Ding, Wang, Gu, Zhao, Wang, and Li]{spatialvla}
D.~Qu, H.~Song, Q.~Chen, Y.~Yao, X.~Ye, Y.~Ding, Z.~Wang, J.~Gu, B.~Zhao, D.~Wang, and X.~Li.
\newblock Spatialvla: Exploring spatial representations for visual-language-action model, 2025.
\newblock URL \url{https://arxiv.org/abs/2501.15830}.

\bibitem[Nair et~al.(2023)Nair, Rajeswaran, Kumar, Finn, and Gupta]{R3M}
S.~Nair, A.~Rajeswaran, V.~Kumar, C.~Finn, and A.~Gupta.
\newblock R3m: A universal visual representation for robot manipulation.
\newblock In K.~Liu, D.~Kulic, and J.~Ichnowski, editors, \emph{Proceedings of The 6th Conference on Robot Learning}, volume 205 of \emph{Proceedings of Machine Learning Research}, pages 892--909. PMLR, 14--18 Dec 2023.
\newblock URL \url{https://proceedings.mlr.press/v205/nair23a.html}.

\bibitem[Radosavovic et~al.(2023)Radosavovic, Xiao, James, Abbeel, Malik, and Darrell]{Masked_Visual_Pre-training}
I.~Radosavovic, T.~Xiao, S.~James, P.~Abbeel, J.~Malik, and T.~Darrell.
\newblock Real-world robot learning with masked visual pre-training.
\newblock In K.~Liu, D.~Kulic, and J.~Ichnowski, editors, \emph{Proceedings of The 6th Conference on Robot Learning}, volume 205 of \emph{Proceedings of Machine Learning Research}, pages 416--426. PMLR, 14--18 Dec 2023.
\newblock URL \url{https://proceedings.mlr.press/v205/radosavovic23a.html}.

\bibitem[Ma et~al.(2023)Ma, Sodhani, Jayaraman, Bastani, Kumar, and Zhang]{VIP}
Y.~J. Ma, S.~Sodhani, D.~Jayaraman, O.~Bastani, V.~Kumar, and A.~Zhang.
\newblock {VIP}: Towards universal visual reward and representation via value-implicit pre-training.
\newblock In \emph{The Eleventh International Conference on Learning Representations}, 2023.
\newblock URL \url{https://openreview.net/forum?id=YJ7o2wetJ2}.

\bibitem[Majumdar et~al.(2023)Majumdar, Yadav, Arnaud, Ma, Chen, Silwal, Jain, Berges, Wu, Vakil, Abbeel, Malik, Batra, Lin, Maksymets, Rajeswaran, and Meier]{Where}
A.~Majumdar, K.~Yadav, S.~Arnaud, J.~Ma, C.~Chen, S.~Silwal, A.~Jain, V.-P. Berges, T.~Wu, J.~Vakil, P.~Abbeel, J.~Malik, D.~Batra, Y.~Lin, O.~Maksymets, A.~Rajeswaran, and F.~Meier.
\newblock Where are we in the search for an artificial visual cortex for embodied intelligence?
\newblock In A.~Oh, T.~Naumann, A.~Globerson, K.~Saenko, M.~Hardt, and S.~Levine, editors, \emph{Advances in Neural Information Processing Systems}, volume~36, pages 655--677. Curran Associates, Inc., 2023.
\newblock URL \url{https://proceedings.neurips.cc/paper_files/paper/2023/file/022ca1bed6b574b962c48a2856eb207b-Paper-Conference.pdf}.

\bibitem[Du et~al.(2023)Du, Yang, Dai, Dai, Nachum, Tenenbaum, Schuurmans, and Abbeel]{UniPi}
Y.~Du, M.~Yang, B.~Dai, H.~Dai, O.~Nachum, J.~B. Tenenbaum, D.~Schuurmans, and P.~Abbeel.
\newblock Learning universal policies via text-guided video generation, 2023.
\newblock URL \url{https://arxiv.org/abs/2302.00111}.

\bibitem[Bharadhwaj et~al.(2024)Bharadhwaj, Dwibedi, Gupta, Tulsiani, Doersch, Xiao, Shah, Xia, Sadigh, and Kirmani]{Gen2Act}
H.~Bharadhwaj, D.~Dwibedi, A.~Gupta, S.~Tulsiani, C.~Doersch, T.~Xiao, D.~Shah, F.~Xia, D.~Sadigh, and S.~Kirmani.
\newblock Gen2act: Human video generation in novel scenarios enables generalizable robot manipulation, 2024.
\newblock URL \url{https://arxiv.org/abs/2409.16283}.

\bibitem[Xu et~al.(2025)Xu, Qiu, and She]{VILP}
Z.~Xu, Q.~Qiu, and Y.~She.
\newblock Vilp: Imitation learning with latent video planning.
\newblock \emph{IEEE Robotics and Automation Letters}, 10\penalty0 (4):\penalty0 3350--3357, 2025.
\newblock \doi{10.1109/LRA.2025.3542317}.

\bibitem[Xu et~al.(2024)Xu, Xu, Xu, Chi, Wetzstein, Veloso, and Song]{Flow2Act}
M.~Xu, Z.~Xu, Y.~Xu, C.~Chi, G.~Wetzstein, M.~Veloso, and S.~Song.
\newblock Flow as the cross-domain manipulation interface.
\newblock In \emph{8th Annual Conference on Robot Learning}, 2024.
\newblock URL \url{https://openreview.net/forum?id=cNI0ZkK1yC}.

\bibitem[Wen et~al.(2024)Wen, Lin, So, Chen, Dou, Gao, and Abbeel]{ATM}
C.~Wen, X.~Lin, J.~So, K.~Chen, Q.~Dou, Y.~Gao, and P.~Abbeel.
\newblock Any-point trajectory modeling for policy learning, 2024.
\newblock URL \url{https://arxiv.org/abs/2401.00025}.

\bibitem[Bharadhwaj et~al.(2024)Bharadhwaj, Mottaghi, Gupta, and Tulsiani]{Track2Act}
H.~Bharadhwaj, R.~Mottaghi, A.~Gupta, and S.~Tulsiani.
\newblock Track2act: Predicting point tracks from internet videos enables generalizable robot manipulation, 2024.
\newblock URL \url{https://arxiv.org/abs/2405.01527}.

\bibitem[Haldar and Pinto(2025)]{PointPolicy}
S.~Haldar and L.~Pinto.
\newblock Point policy: Unifying observations and actions with key points for robot manipulation, 2025.
\newblock URL \url{https://arxiv.org/abs/2502.20391}.

\bibitem[Ren et~al.(2025)Ren, Sundaresan, Sadigh, Choudhury, and Bohg]{MotionTracks}
J.~Ren, P.~Sundaresan, D.~Sadigh, S.~Choudhury, and J.~Bohg.
\newblock Motion tracks: A unified representation for human-robot transfer in few-shot imitation learning, 2025.
\newblock URL \url{https://arxiv.org/abs/2501.06994}.

\bibitem[Kareer et~al.(2024)Kareer, Patel, Punamiya, Mathur, Cheng, Wang, Hoffman, and Xu]{EgoMimic}
S.~Kareer, D.~Patel, R.~Punamiya, P.~Mathur, S.~Cheng, C.~Wang, J.~Hoffman, and D.~Xu.
\newblock Egomimic: Scaling imitation learning via egocentric video, 2024.
\newblock URL \url{https://arxiv.org/abs/2410.24221}.

\bibitem[Bharadhwaj et~al.(2023)Bharadhwaj, Gupta, Tulsiani, and Kumar]{ZeroShot}
H.~Bharadhwaj, A.~Gupta, S.~Tulsiani, and V.~Kumar.
\newblock Zero-shot robot manipulation from passive human videos, 2023.
\newblock URL \url{https://arxiv.org/abs/2302.02011}.

\bibitem[Singh et~al.(2024)Singh, Loquercio, Sferrazza, Wu, Qi, Abbeel, and Malik]{HandObject}
H.~G. Singh, A.~Loquercio, C.~Sferrazza, J.~Wu, H.~Qi, P.~Abbeel, and J.~Malik.
\newblock Hand-object interaction pretraining from videos.
\newblock \emph{CoRR}, abs/2409.08273, 2024.
\newblock URL \url{https://doi.org/10.48550/arXiv.2409.08273}.

\bibitem[Black et~al.(2024)Black, Nakamoto, Atreya, Walke, Finn, Kumar, and Levine]{susie}
K.~Black, M.~Nakamoto, P.~Atreya, H.~R. Walke, C.~Finn, A.~Kumar, and S.~Levine.
\newblock Zero-shot robotic manipulation with pre-trained image-editing diffusion models.
\newblock In \emph{The Twelfth International Conference on Learning Representations}, 2024.
\newblock URL \url{https://openreview.net/forum?id=c0chJTSbci}.

\bibitem[Shridhar et~al.(2024)Shridhar, Lo, and James]{genima}
M.~Shridhar, Y.~L. Lo, and S.~James.
\newblock Generative image as action models.
\newblock In \emph{8th Annual Conference on Robot Learning}, 2024.
\newblock URL \url{https://openreview.net/forum?id=cocHfT7CEs}.

\bibitem[Chen et~al.(2023)Chen, Wang, Beyer, Kolesnikov, Wu, Voigtlaender, Mustafa, Goodman, Alabdulmohsin, Padlewski, Salz, Xiong, Vlasic, Pavetic, Rong, Yu, Keysers, Zhai, and Soricut]{pali3}
X.~Chen, X.~Wang, L.~Beyer, A.~Kolesnikov, J.~Wu, P.~Voigtlaender, B.~Mustafa, S.~Goodman, I.~Alabdulmohsin, P.~Padlewski, D.~Salz, X.~Xiong, D.~Vlasic, F.~Pavetic, K.~Rong, T.~Yu, D.~Keysers, X.~Zhai, and R.~Soricut.
\newblock Pali-3 vision language models: Smaller, faster, stronger, 2023.
\newblock URL \url{https://arxiv.org/abs/2310.09199}.

\bibitem[Beyer et~al.(2024)Beyer, Steiner, Pinto, Kolesnikov, Wang, Salz, Neumann, Alabdulmohsin, Tschannen, Bugliarello, Unterthiner, Keysers, Koppula, Liu, Grycner, Gritsenko, Houlsby, Kumar, Rong, Eisenschlos, Kabra, Bauer, Bosnjak, Chen, Minderer, Voigtlaender, Bica, Balazevic, Puigcerver, Papalampidi, Hénaff, Xiong, Soricut, Harmsen, and Zhai]{Paligemma}
L.~Beyer, A.~Steiner, A.~S. Pinto, A.~Kolesnikov, X.~Wang, D.~Salz, M.~Neumann, I.~Alabdulmohsin, M.~Tschannen, E.~Bugliarello, T.~Unterthiner, D.~Keysers, S.~Koppula, F.~Liu, A.~Grycner, A.~A. Gritsenko, N.~Houlsby, M.~Kumar, K.~Rong, J.~Eisenschlos, R.~Kabra, M.~Bauer, M.~Bosnjak, X.~Chen, M.~Minderer, P.~Voigtlaender, I.~Bica, I.~Balazevic, J.~Puigcerver, P.~Papalampidi, O.~J. Hénaff, X.~Xiong, R.~Soricut, J.~Harmsen, and X.~Zhai.
\newblock Paligemma: A versatile 3b vlm for transfer.
\newblock \emph{CoRR}, abs/2407.07726, 2024.
\newblock URL \url{https://doi.org/10.48550/arXiv.2407.07726}.

\bibitem[Team et~al.(2024)Team, Riviere, Pathak, Sessa, Hardin, Bhupatiraju, Hussenot, Mesnard, Shahriari, Ramé, Ferret, Liu, Tafti, Friesen, Casbon, Ramos, Kumar, Lan, Jerome, Tsitsulin, Vieillard, Stanczyk, Girgin, Momchev, Hoffman, Thakoor, Grill, Neyshabur, Bachem, Walton, Severyn, Parrish, Ahmad, Hutchison, Abdagic, Carl, Shen, Brock, Coenen, Laforge, Paterson, Bastian, Piot, Wu, Royal, Chen, Kumar, Perry, Welty, Choquette-Choo, Sinopalnikov, Weinberger, Vijaykumar, Rogozińska, Herbison, Bandy, Wang, Noland, Moreira, Senter, Eltyshev, Visin, Rasskin, Wei, Cameron, Martins, Hashemi, Klimczak-Plucińska, Batra, Dhand, Nardini, Mein, Zhou, Svensson, Stanway, Chan, Zhou, Carrasqueira, Iljazi, Becker, Fernandez, van Amersfoort, Gordon, Lipschultz, Newlan, yeong Ji, Mohamed, Badola, Black, Millican, McDonell, Nguyen, Sodhia, Greene, Sjoesund, Usui, Sifre, Heuermann, Lago, McNealus, Soares, Kilpatrick, Dixon, Martins, Reid, Singh, Iverson, Görner, Velloso, Wirth, Davidow, Miller, Rahtz, Watson, Risdal, Kazemi, Moynihan, Zhang, Kahng, Park, Rahman, Khatwani, Dao, Bardoliwalla, Devanathan, Dumai, Chauhan, Wahltinez, Botarda, Barnes, Barham, Michel, Jin, Georgiev, Culliton, Kuppala, Comanescu, Merhej, Jana, Rokni, Agarwal, Mullins, Saadat, Carthy, Cogan, Perrin, Arnold, Krause, Dai, Garg, Sheth, Ronstrom, Chan, Jordan, Yu, Eccles, Hennigan, Kocisky, Doshi, Jain, Yadav, Meshram, Dharmadhikari, Barkley, Wei, Ye, Han, Kwon, Xu, Shen, Gong, Wei, Cotruta, Kirk, Rao, Giang, Peran, Warkentin, Collins, Barral, Ghahramani, Hadsell, Sculley, Banks, Dragan, Petrov, Vinyals, Dean, Hassabis, Kavukcuoglu, Farabet, Buchatskaya, Borgeaud, Fiedel, Joulin, Kenealy, Dadashi, and Andreev]{gemma2}
G.~Team, M.~Riviere, S.~Pathak, P.~G. Sessa, C.~Hardin, S.~Bhupatiraju, L.~Hussenot, T.~Mesnard, B.~Shahriari, A.~Ramé, J.~Ferret, P.~Liu, P.~Tafti, A.~Friesen, M.~Casbon, S.~Ramos, R.~Kumar, C.~L. Lan, S.~Jerome, A.~Tsitsulin, N.~Vieillard, P.~Stanczyk, S.~Girgin, N.~Momchev, M.~Hoffman, S.~Thakoor, J.-B. Grill, B.~Neyshabur, O.~Bachem, A.~Walton, A.~Severyn, A.~Parrish, A.~Ahmad, A.~Hutchison, A.~Abdagic, A.~Carl, A.~Shen, A.~Brock, A.~Coenen, A.~Laforge, A.~Paterson, B.~Bastian, B.~Piot, B.~Wu, B.~Royal, C.~Chen, C.~Kumar, C.~Perry, C.~Welty, C.~A. Choquette-Choo, D.~Sinopalnikov, D.~Weinberger, D.~Vijaykumar, D.~Rogozińska, D.~Herbison, E.~Bandy, E.~Wang, E.~Noland, E.~Moreira, E.~Senter, E.~Eltyshev, F.~Visin, G.~Rasskin, G.~Wei, G.~Cameron, G.~Martins, H.~Hashemi, H.~Klimczak-Plucińska, H.~Batra, H.~Dhand, I.~Nardini, J.~Mein, J.~Zhou, J.~Svensson, J.~Stanway, J.~Chan, J.~P. Zhou, J.~Carrasqueira, J.~Iljazi, J.~Becker, J.~Fernandez, J.~van Amersfoort, J.~Gordon, J.~Lipschultz, J.~Newlan, J.~yeong Ji, K.~Mohamed, K.~Badola, K.~Black, K.~Millican, K.~McDonell, K.~Nguyen, K.~Sodhia, K.~Greene, L.~L. Sjoesund, L.~Usui, L.~Sifre, L.~Heuermann, L.~Lago, L.~McNealus, L.~B. Soares, L.~Kilpatrick, L.~Dixon, L.~Martins, M.~Reid, M.~Singh, M.~Iverson, M.~Görner, M.~Velloso, M.~Wirth, M.~Davidow, M.~Miller, M.~Rahtz, M.~Watson, M.~Risdal, M.~Kazemi, M.~Moynihan, M.~Zhang, M.~Kahng, M.~Park, M.~Rahman, M.~Khatwani, N.~Dao, N.~Bardoliwalla, N.~Devanathan, N.~Dumai, N.~Chauhan, O.~Wahltinez, P.~Botarda, P.~Barnes, P.~Barham, P.~Michel, P.~Jin, P.~Georgiev, P.~Culliton, P.~Kuppala, R.~Comanescu, R.~Merhej, R.~Jana, R.~A. Rokni, R.~Agarwal, R.~Mullins, S.~Saadat, S.~M. Carthy, S.~Cogan, S.~Perrin, S.~M.~R. Arnold, S.~Krause, S.~Dai, S.~Garg, S.~Sheth, S.~Ronstrom, S.~Chan, T.~Jordan, T.~Yu, T.~Eccles, T.~Hennigan, T.~Kocisky, T.~Doshi, V.~Jain, V.~Yadav, V.~Meshram, V.~Dharmadhikari, W.~Barkley, W.~Wei, W.~Ye, W.~Han, W.~Kwon, X.~Xu, Z.~Shen, Z.~Gong, Z.~Wei, V.~Cotruta, P.~Kirk, A.~Rao, M.~Giang, L.~Peran, T.~Warkentin, E.~Collins, J.~Barral, Z.~Ghahramani, R.~Hadsell, D.~Sculley, J.~Banks, A.~Dragan, S.~Petrov, O.~Vinyals, J.~Dean, D.~Hassabis, K.~Kavukcuoglu, C.~Farabet, E.~Buchatskaya, S.~Borgeaud, N.~Fiedel, A.~Joulin, K.~Kenealy, R.~Dadashi, and A.~Andreev.
\newblock Gemma 2: Improving open language models at a practical size, 2024.
\newblock URL \url{https://arxiv.org/abs/2408.00118}.

\bibitem[Zhai et~al.(2023)Zhai, Mustafa, Kolesnikov, and Beyer]{SigLIP}
X.~Zhai, B.~Mustafa, A.~Kolesnikov, and L.~Beyer.
\newblock Sigmoid loss for language image pre-training.
\newblock In \emph{Proceedings of the IEEE/CVF International Conference on Computer Vision (ICCV)}, pages 11975--11986, October 2023.

\bibitem[Radford et~al.(2021)Radford, Kim, Hallacy, Ramesh, Goh, Agarwal, Sastry, Askell, Mishkin, Clark, Krueger, and Sutskever]{VLM}
A.~Radford, J.~W. Kim, C.~Hallacy, A.~Ramesh, G.~Goh, S.~Agarwal, G.~Sastry, A.~Askell, P.~Mishkin, J.~Clark, G.~Krueger, and I.~Sutskever.
\newblock Learning transferable visual models from natural language supervision.
\newblock In M.~Meila and T.~Zhang, editors, \emph{Proceedings of the 38th International Conference on Machine Learning}, volume 139 of \emph{Proceedings of Machine Learning Research}, pages 8748--8763. PMLR, 18--24 Jul 2021.
\newblock URL \url{https://proceedings.mlr.press/v139/radford21a.html}.

\bibitem[Ahn et~al.(2022)Ahn, Brohan, Brown, Chebotar, Cortes, David, Finn, Fu, Gopalakrishnan, Hausman, Herzog, Ho, Hsu, Ibarz, Ichter, Irpan, Jang, Ruano, Jeffrey, Jesmonth, Joshi, Julian, Kalashnikov, Kuang, Lee, Levine, Lu, Luu, Parada, Pastor, Quiambao, Rao, Rettinghouse, Reyes, Sermanet, Sievers, Tan, Toshev, Vanhoucke, Xia, Xiao, Xu, Xu, Yan, and Zeng]{SayCan}
M.~Ahn, A.~Brohan, N.~Brown, Y.~Chebotar, O.~Cortes, B.~David, C.~Finn, C.~Fu, K.~Gopalakrishnan, K.~Hausman, A.~Herzog, D.~Ho, J.~Hsu, J.~Ibarz, B.~Ichter, A.~Irpan, E.~Jang, R.~J. Ruano, K.~Jeffrey, S.~Jesmonth, N.~J. Joshi, R.~Julian, D.~Kalashnikov, Y.~Kuang, K.-H. Lee, S.~Levine, Y.~Lu, L.~Luu, C.~Parada, P.~Pastor, J.~Quiambao, K.~Rao, J.~Rettinghouse, D.~Reyes, P.~Sermanet, N.~Sievers, C.~Tan, A.~Toshev, V.~Vanhoucke, F.~Xia, T.~Xiao, P.~Xu, S.~Xu, M.~Yan, and A.~Zeng.
\newblock Do as i can, not as i say: Grounding language in robotic affordances, 2022.
\newblock URL \url{https://arxiv.org/abs/2204.01691}.

\bibitem[Huang et~al.(2022)Huang, Abbeel, Pathak, and Mordatch]{huang2022language}
W.~Huang, P.~Abbeel, D.~Pathak, and I.~Mordatch.
\newblock Language models as zero-shot planners: Extracting actionable knowledge for embodied agents.
\newblock In K.~Chaudhuri, S.~Jegelka, L.~Song, C.~Szepesvari, G.~Niu, and S.~Sabato, editors, \emph{Proceedings of the 39th International Conference on Machine Learning}, volume 162 of \emph{Proceedings of Machine Learning Research}, pages 9118--9147. PMLR, 17--23 Jul 2022.
\newblock URL \url{https://proceedings.mlr.press/v162/huang22a.html}.

\bibitem[Singh et~al.(2023)Singh, Blukis, Mousavian, Goyal, Xu, Tremblay, Fox, Thomason, and Garg]{ProgPrompt}
I.~Singh, V.~Blukis, A.~Mousavian, A.~Goyal, D.~Xu, J.~Tremblay, D.~Fox, J.~Thomason, and A.~Garg.
\newblock Progprompt: Generating situated robot task plans using large language models.
\newblock In \emph{2023 IEEE International Conference on Robotics and Automation (ICRA)}, pages 11523--11530, 2023.
\newblock \doi{10.1109/ICRA48891.2023.10161317}.

\bibitem[Driess et~al.(2023)Driess, Xia, Sajjadi, Lynch, Chowdhery, Ichter, Wahid, Tompson, Vuong, Yu, Huang, Chebotar, Sermanet, Duckworth, Levine, Vanhoucke, Hausman, Toussaint, Greff, Zeng, Mordatch, and Florence]{Palme}
D.~Driess, F.~Xia, M.~S.~M. Sajjadi, C.~Lynch, A.~Chowdhery, B.~Ichter, A.~Wahid, J.~Tompson, Q.~Vuong, T.~Yu, W.~Huang, Y.~Chebotar, P.~Sermanet, D.~Duckworth, S.~Levine, V.~Vanhoucke, K.~Hausman, M.~Toussaint, K.~Greff, A.~Zeng, I.~Mordatch, and P.~Florence.
\newblock Palm-e: An embodied multimodal language model.
\newblock In \emph{arXiv preprint arXiv:2303.03378}, 2023.

\bibitem[Karamcheti et~al.(2024)Karamcheti, Nair, Balakrishna, Liang, Kollar, and Sadigh]{Prismatic}
S.~Karamcheti, S.~Nair, A.~Balakrishna, P.~Liang, T.~Kollar, and D.~Sadigh.
\newblock Prismatic {VLM}s: Investigating the design space of visually-conditioned language models.
\newblock In \emph{Forty-first International Conference on Machine Learning}, 2024.
\newblock URL \url{https://openreview.net/forum?id=6FXtu8clyp}.

\bibitem[Zhen et~al.(2024)Zhen, Qiu, Chen, Yang, Yan, Du, Hong, and Gan]{3DVLA}
H.~Zhen, X.~Qiu, P.~Chen, J.~Yang, X.~Yan, Y.~Du, Y.~Hong, and C.~Gan.
\newblock 3{D}-{VLA}: A 3{D} vision-language-action generative world model.
\newblock In R.~Salakhutdinov, Z.~Kolter, K.~Heller, A.~Weller, N.~Oliver, J.~Scarlett, and F.~Berkenkamp, editors, \emph{Proceedings of the 41st International Conference on Machine Learning}, volume 235 of \emph{Proceedings of Machine Learning Research}, pages 61229--61245. PMLR, 21--27 Jul 2024.
\newblock URL \url{https://proceedings.mlr.press/v235/zhen24a.html}.

\bibitem[He et~al.(2022)He, Chen, Xie, Li, Doll\'ar, and Girshick]{He_2022_CVPR}
K.~He, X.~Chen, S.~Xie, Y.~Li, P.~Doll\'ar, and R.~Girshick.
\newblock Masked autoencoders are scalable vision learners.
\newblock In \emph{Proceedings of the IEEE/CVF Conference on Computer Vision and Pattern Recognition (CVPR)}, pages 16000--16009, June 2022.

\bibitem[Niu et~al.(2025)Niu, Sharma, Xue, Biamby, Zhang, Ji, Darrell, and Herzig]{Arm4r}
D.~Niu, Y.~Sharma, H.~Xue, G.~Biamby, J.~Zhang, Z.~Ji, T.~Darrell, and R.~Herzig.
\newblock Pre-training auto-regressive robotic models with 4d representations, 2025.
\newblock URL \url{https://arxiv.org/abs/2502.13142}.

\bibitem[Ye et~al.(2025)Ye, Jang, Jeon, Joo, Yang, Peng, Mandlekar, Tan, Chao, Lin, Liden, Lee, Gao, Zettlemoyer, Fox, and Seo]{LAPA}
S.~Ye, J.~Jang, B.~Jeon, S.~J. Joo, J.~Yang, B.~Peng, A.~Mandlekar, R.~Tan, Y.-W. Chao, B.~Y. Lin, L.~Liden, K.~Lee, J.~Gao, L.~Zettlemoyer, D.~Fox, and M.~Seo.
\newblock Latent action pretraining from videos.
\newblock In \emph{The Thirteenth International Conference on Learning Representations}, 2025.
\newblock URL \url{https://openreview.net/forum?id=VYOe2eBQeh}.

\bibitem[Liu et~al.(2024)Liu, Zeng, Ren, Li, Zhang, Yang, Jiang, Li, Yang, Su, Zhu, and Zhang]{GroundingDINO}
S.~Liu, Z.~Zeng, T.~Ren, F.~Li, H.~Zhang, J.~Yang, Q.~Jiang, C.~Li, J.~Yang, H.~Su, J.~Zhu, and L.~Zhang.
\newblock Grounding dino: Marrying dino with grounded pre-training for open-set object detection, 2024.
\newblock URL \url{https://arxiv.org/abs/2303.05499}.

\bibitem[Ravi et~al.(2024)Ravi, Gabeur, Hu, Hu, Ryali, Ma, Khedr, Rädle, Rolland, Gustafson, Mintun, Pan, Alwala, Carion, Wu, Girshick, Dollár, and Feichtenhofer]{SAM2}
N.~Ravi, V.~Gabeur, Y.-T. Hu, R.~Hu, C.~Ryali, T.~Ma, H.~Khedr, R.~Rädle, C.~Rolland, L.~Gustafson, E.~Mintun, J.~Pan, K.~V. Alwala, N.~Carion, C.-Y. Wu, R.~Girshick, P.~Dollár, and C.~Feichtenhofer.
\newblock Sam 2: Segment anything in images and videos, 2024.
\newblock URL \url{https://arxiv.org/abs/2408.00714}.

\bibitem[Doersch et~al.(2024)Doersch, Luc, Yang, Gokay, Koppula, Gupta, Heyward, Rocco, Goroshin, Carreira, and Zisserman]{BootsTAP}
C.~Doersch, P.~Luc, Y.~Yang, D.~Gokay, S.~Koppula, A.~Gupta, J.~Heyward, I.~Rocco, R.~Goroshin, J.~a. Carreira, and A.~Zisserman.
\newblock Bootstap: Bootstrapped training for tracking-any-point.
\newblock In \emph{Proceedings of the Asian Conference on Computer Vision (ACCV)}, pages 3257--3274, December 2024.

\bibitem[Wang et~al.(2024)Wang, Xu, Dai, Xiang, Deng, Tong, and Yang]{MoGE}
R.~Wang, S.~Xu, C.~Dai, J.~Xiang, Y.~Deng, X.~Tong, and J.~Yang.
\newblock Moge: Unlocking accurate monocular geometry estimation for open-domain images with optimal training supervision, 2024.
\newblock URL \url{https://arxiv.org/abs/2410.19115}.

\bibitem[Liang et~al.(2024)Liang, Yu, Luo, Iyer, Dong, Zhou, Ghosh, Lewis, tau Yih, Zettlemoyer, and Lin]{MixtureOfTransformers}
W.~Liang, L.~Yu, L.~Luo, S.~Iyer, N.~Dong, C.~Zhou, G.~Ghosh, M.~Lewis, W.~tau Yih, L.~Zettlemoyer, and X.~V. Lin.
\newblock Mixture-of-transformers: A sparse and scalable architecture for multi-modal foundation models, 2024.
\newblock URL \url{https://arxiv.org/abs/2411.04996}.

\bibitem[Vaswani et~al.(2017)Vaswani, Shazeer, Parmar, Uszkoreit, Jones, Gomez, Kaiser, and Polosukhin]{Transformer}
A.~Vaswani, N.~Shazeer, N.~Parmar, J.~Uszkoreit, L.~Jones, A.~N. Gomez, L.~u. Kaiser, and I.~Polosukhin.
\newblock Attention is all you need.
\newblock In I.~Guyon, U.~V. Luxburg, S.~Bengio, H.~Wallach, R.~Fergus, S.~Vishwanathan, and R.~Garnett, editors, \emph{Advances in Neural Information Processing Systems}, volume~30. Curran Associates, Inc., 2017.
\newblock URL \url{https://proceedings.neurips.cc/paper_files/paper/2017/file/3f5ee243547dee91fbd053c1c4a845aa-Paper.pdf}.

\bibitem[Lipman et~al.(2023)Lipman, Chen, Ben-Hamu, Nickel, and Le]{FlowMatching}
Y.~Lipman, R.~T.~Q. Chen, H.~Ben-Hamu, M.~Nickel, and M.~Le.
\newblock Flow matching for generative modeling.
\newblock In \emph{The Eleventh International Conference on Learning Representations}, 2023.
\newblock URL \url{https://openreview.net/forum?id=PqvMRDCJT9t}.

\bibitem[Mesnard et~al.(2024)Mesnard, Hardin, Dadashi, Bhupatiraju, Pathak, Sifre, Rivière, Kale, Love, Tafti, Hussenot, Chowdhery, Roberts, Barua, Botev, Castro-Ros, Slone, Héliou, Tacchetti, Bulanova, Paterson, Tsai, Shahriari, Lan, Choquette-Choo, Crepy, Cer, Ippolito, Reid, Buchatskaya, Ni, Noland, Yan, Tucker, Muraru, Rozhdestvenskiy, Michalewski, Tenney, Grishchenko, Austin, Keeling, Labanowski, Lespiau, Stanway, Brennan, Chen, Ferret, Chiu, and et~al.]{gemma}
T.~Mesnard, C.~Hardin, R.~Dadashi, S.~Bhupatiraju, S.~Pathak, L.~Sifre, M.~Rivière, M.~S. Kale, J.~Love, P.~Tafti, L.~Hussenot, A.~Chowdhery, A.~Roberts, A.~Barua, A.~Botev, A.~Castro-Ros, A.~Slone, A.~Héliou, A.~Tacchetti, A.~Bulanova, A.~Paterson, B.~Tsai, B.~Shahriari, C.~L. Lan, C.~A. Choquette-Choo, C.~Crepy, D.~Cer, D.~Ippolito, D.~Reid, E.~Buchatskaya, E.~Ni, E.~Noland, G.~Yan, G.~Tucker, G.-C. Muraru, G.~Rozhdestvenskiy, H.~Michalewski, I.~Tenney, I.~Grishchenko, J.~Austin, J.~Keeling, J.~Labanowski, J.-B. Lespiau, J.~Stanway, J.~Brennan, J.~Chen, J.~Ferret, J.~Chiu, and et~al.
\newblock Gemma: Open models based on gemini research and technology.
\newblock \emph{CoRR}, abs/2403.08295, 2024.
\newblock URL \url{https://doi.org/10.48550/arXiv.2403.08295}.

\bibitem[Fang et~al.(2023)Fang, Fang, Tang, Liu, Wang, Wang, Zhu, and Lu]{rh20t}
H.-S. Fang, H.~Fang, Z.~Tang, J.~Liu, C.~Wang, J.~Wang, H.~Zhu, and C.~Lu.
\newblock Rh20t: A comprehensive robotic dataset for learning diverse skills in one-shot, 2023.
\newblock URL \url{https://arxiv.org/abs/2307.00595}.

\bibitem[Walke et~al.(2023)Walke, Black, Zhao, Vuong, Zheng, Hansen-Estruch, He, Myers, Kim, Du, Lee, Fang, Finn, and Levine]{BridgeV2}
H.~R. Walke, K.~Black, T.~Z. Zhao, Q.~Vuong, C.~Zheng, P.~Hansen-Estruch, A.~W. He, V.~Myers, M.~J. Kim, M.~Du, A.~Lee, K.~Fang, C.~Finn, and S.~Levine.
\newblock Bridgedata v2: A dataset for robot learning at scale.
\newblock In J.~Tan, M.~Toussaint, and K.~Darvish, editors, \emph{Proceedings of The 7th Conference on Robot Learning}, volume 229 of \emph{Proceedings of Machine Learning Research}, pages 1723--1736. PMLR, 06--09 Nov 2023.
\newblock URL \url{https://proceedings.mlr.press/v229/walke23a.html}.

\bibitem[Brohan et~al.(2022)Brohan, Brown, Carbajal, Chebotar, Dabis, Finn, Gopalakrishnan, Hausman, Herzog, Hsu, et~al.]{brohan2022rt}
A.~Brohan, N.~Brown, J.~Carbajal, Y.~Chebotar, J.~Dabis, C.~Finn, K.~Gopalakrishnan, K.~Hausman, A.~Herzog, J.~Hsu, et~al.
\newblock Rt-1: Robotics transformer for real-world control at scale.
\newblock \emph{arXiv preprint arXiv:2212.06817}, 2022.

\bibitem[Li et~al.(2024)Li, Hsu, Gu, Pertsch, Mees, Walke, Fu, Lunawat, Sieh, Kirmani, Levine, Wu, Finn, Su, Vuong, and Xiao]{simpler}
X.~Li, K.~Hsu, J.~Gu, K.~Pertsch, O.~Mees, H.~R. Walke, C.~Fu, I.~Lunawat, I.~Sieh, S.~Kirmani, S.~Levine, J.~Wu, C.~Finn, H.~Su, Q.~Vuong, and T.~Xiao.
\newblock Evaluating real-world robot manipulation policies in simulation.
\newblock \emph{arXiv preprint arXiv:2405.05941}, 2024.

\bibitem[Damen et~al.(2022)Damen, Doughty, Farinella, Furnari, Ma, Kazakos, Moltisanti, Munro, Perrett, Price, and Wray]{EpicKitchen}
D.~Damen, H.~Doughty, G.~M. Farinella, A.~Furnari, J.~Ma, E.~Kazakos, D.~Moltisanti, J.~Munro, T.~Perrett, W.~Price, and M.~Wray.
\newblock Rescaling egocentric vision: Collection, pipeline and challenges for epic-kitchens-100.
\newblock \emph{International Journal of Computer Vision (IJCV)}, 130:\penalty0 33–55, 2022.
\newblock URL \url{https://doi.org/10.1007/s11263-021-01531-2}.

\bibitem[Grauman et~al.(2022)Grauman, Westbury, Byrne, Chavis, Furnari, Girdhar, Hamburger, Jiang, Liu, Liu, Martin, Nagarajan, Radosavovic, Ramakrishnan, Ryan, Sharma, Wray, Xu, Xu, Zhao, Bansal, Batra, Cartillier, Crane, Do, Doulaty, Erapalli, Feichtenhofer, Fragomeni, Fu, Gebreselasie, Gonzalez, Hillis, Huang, Huang, Jia, Khoo, Kolar, Kottur, Kumar, Landini, Li, Li, Li, Mangalam, Modhugu, Munro, Murrell, Nishiyasu, Price, Puentes, Ramazanova, Sari, Somasundaram, Southerland, Sugano, Tao, Vo, Wang, Wu, Yagi, Zhao, Zhu, Arbelaez, Crandall, Damen, Farinella, Fuegen, Ghanem, Ithapu, Jawahar, Joo, Kitani, Li, Newcombe, Oliva, Park, Rehg, Sato, Shi, Shou, Torralba, Torresani, Yan, and Malik]{Ego4D}
K.~Grauman, A.~Westbury, E.~Byrne, Z.~Chavis, A.~Furnari, R.~Girdhar, J.~Hamburger, H.~Jiang, M.~Liu, X.~Liu, M.~Martin, T.~Nagarajan, I.~Radosavovic, S.~K. Ramakrishnan, F.~Ryan, J.~Sharma, M.~Wray, M.~Xu, E.~Z. Xu, C.~Zhao, S.~Bansal, D.~Batra, V.~Cartillier, S.~Crane, T.~Do, M.~Doulaty, A.~Erapalli, C.~Feichtenhofer, A.~Fragomeni, Q.~Fu, A.~Gebreselasie, C.~Gonzalez, J.~Hillis, X.~Huang, Y.~Huang, W.~Jia, W.~Khoo, J.~Kolar, S.~Kottur, A.~Kumar, F.~Landini, C.~Li, Y.~Li, Z.~Li, K.~Mangalam, R.~Modhugu, J.~Munro, T.~Murrell, T.~Nishiyasu, W.~Price, P.~R. Puentes, M.~Ramazanova, L.~Sari, K.~Somasundaram, A.~Southerland, Y.~Sugano, R.~Tao, M.~Vo, Y.~Wang, X.~Wu, T.~Yagi, Z.~Zhao, Y.~Zhu, P.~Arbelaez, D.~Crandall, D.~Damen, G.~M. Farinella, C.~Fuegen, B.~Ghanem, V.~K. Ithapu, C.~V. Jawahar, H.~Joo, K.~Kitani, H.~Li, R.~Newcombe, A.~Oliva, H.~S. Park, J.~M. Rehg, Y.~Sato, J.~Shi, M.~Z. Shou, A.~Torralba, L.~Torresani, M.~Yan, and J.~Malik.
\newblock Ego4d: Around the world in 3,000 hours of egocentric video, 2022.
\newblock URL \url{https://arxiv.org/abs/2110.07058}.

\bibitem[Zhang et~al.(2024)Zhang, Herrmann, Hur, Jampani, Darrell, Cole, Sun, and Yang]{Monster}
J.~Zhang, C.~Herrmann, J.~Hur, V.~Jampani, T.~Darrell, F.~Cole, D.~Sun, and M.-H. Yang.
\newblock Monst3r: A simple approach for estimating geometry in the presence of motion, 2024.
\newblock URL \url{https://arxiv.org/abs/2410.03825}.

\end{thebibliography}
